\documentclass[preprint,sort&compress,9pt,a4paper]{article}
\usepackage{graphicx}
\usepackage{amssymb}
\usepackage[tbtags]{amsmath}
\usepackage[marginal]{footmisc}
\usepackage{bm}
\usepackage{changepage}
\usepackage{caption,subcaption}
\usepackage[colorlinks, linkcolor=blue, anchorcolor=blue,citecolor=blue]{hyperref}
\usepackage{amsthm}
\usepackage{wrapfig}
\usepackage{float}
\usepackage{algorithm}
\usepackage{algorithmic}
\usepackage{authblk}
\usepackage[margin=1in]{geometry}
\allowdisplaybreaks[4]

\title{Transfer Learning-Based Label Proportions Method with Data of Uncertainty}
\author[a]{Yanshan Xiao}
\author[a]{HuaiPei Wang}
\author[b]{Bo Liu \thanks{Corresponding author: csbliu@189.cn}}
\affil[a]{School of Computers, Guangdong University of Technology, Guangzhou, China}
\affil[b]{School of Automation, Guangdong University of Technology, Guangzhou, China}

\date{}
\begin{document}
\maketitle
\begin{abstract}
Learning with label proportions (LLP), which is a learning task that only provides unlabeled data in bags and each bag's label proportion, has widespread successful applications in practice. However, most of the existing LLP methods don't consider the knowledge transfer  for uncertain data. This paper presents a transfer learning-based approach for the problem of learning with label proportions(TL-LLP) to transfer knowledge from source task to target task where both the source and target tasks contain uncertain data. Our approach first formulates objective model for the uncertain data and deals with transfer learning at the same time, and then proposes an iterative framework to build an accurate classifier for the target task. Extensive experiments have shown that the proposed TL-LLP method can obtain the better accuracies and is less sensitive to noise compared with the existing LLP methods.

\end{abstract}

\section{Introduction}
\begin{figure}[t]
    \centering
    \includegraphics[width=4in]{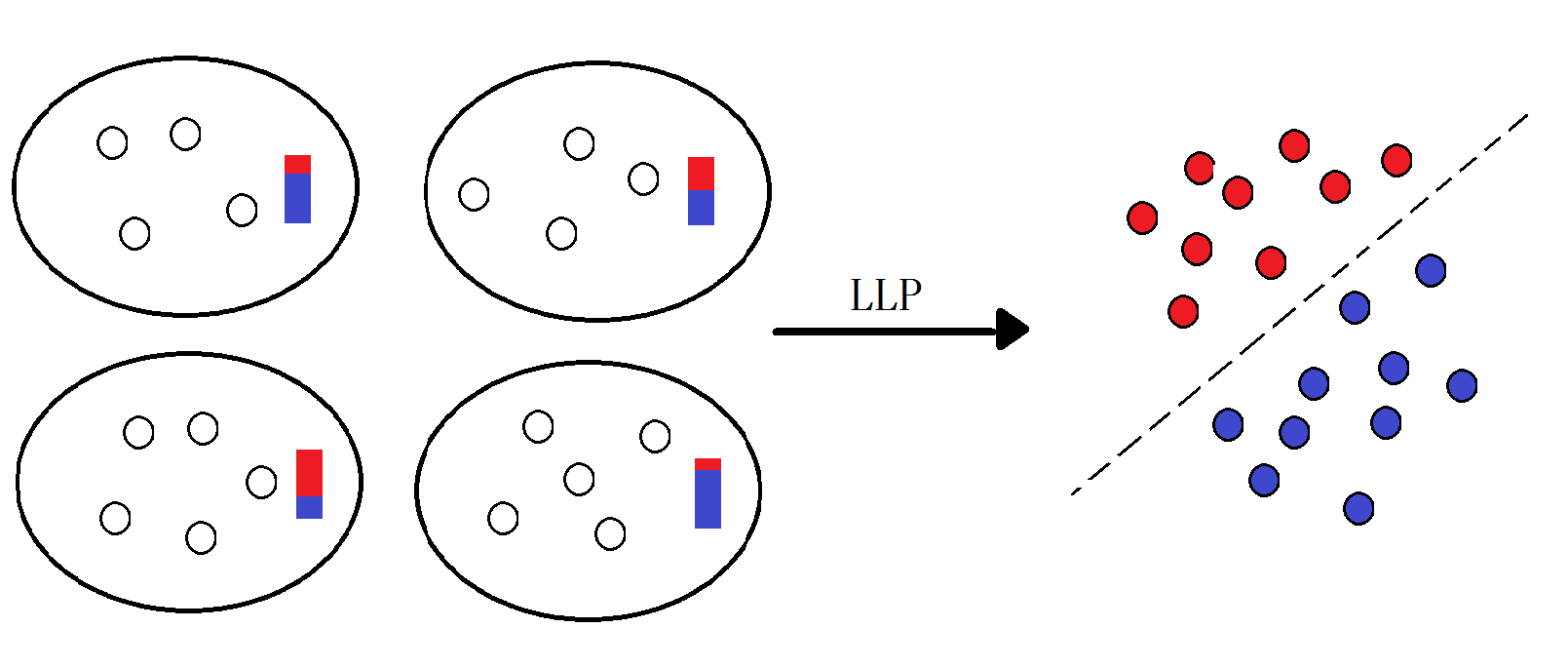}
    \caption{The  binary classification problem of learning with label proportions}
    \label{fig:LLP}
\end{figure}

Learning with label proportions(LLP), which seeks an instance-level classifier merely based on bag-level label proportions, is a new paradigm in machine learning that addresses the classification of instances \cite{kuck2005learning,quadrianto2008estimating,quadrianto2009estimating}. In LLP, we only know the proportions of examples belonging to different classes in each bag; however the labels of the instances are unknown. From the binary classification perspective, the task of LLP is to learn a classifier to classify the unknown label instance as either positive class or negative class. The formulation that learning with label proportions has been first proposed by Kuck et al. in \cite{kuck2005learning}, which can be used for political elections analysis. In the case of politician polls, each candidate may have a group of loyal voters and some swing voters. They may know the vague proportion of votes cast in each district; however, they usually  do not know the vote of each person. Since the candidates have limited resources, they have to analyze political elections and consider which kind of voters they should focus on so as to maximize their interests. To date, LLP has been applied to forecasting revenue \cite{yu2014learning}, image classification \cite{wang2013multi,yu2014modeling}, video event detection \cite{lai2014video}, demographics mining \cite{ardehaly2017mining} and privacy protection \cite{yu2013propto}. Figure \ref{fig:LLP} illustrates the binary classification  problem in LLP. The black circle ``$\circ$" denotes the unlabeled instance. In each bag, the red and blue rectangles represent the amount of negative and positive instances respectively, and we are assumed to know the vague proportion of positive class size and negative class size in advance. On the right, the red and blue samples denote the negative and positive instances, and the dotted line denotes the classifier trained by the label proportions and instances without labels.

Depend on the models of the learning methods, the previous works to LLP can be classified into two broad categories. (1) The approaches for LLP based on support vector machine(SVM) technology \cite{yu2013propto,wang2015linear}, where large-margin framework is proposed to solve LLP problem. For example, Wang et al.\cite{wang2015linear} propose a classification model based on twin SVM, which is in a large-margin framework. (2) Some other methods are proposed to deal with LLP problem based on probabilistic models \cite{hernandez2011learning,fan2014learning}. For example, Fan et al.\cite{fan2014learning} propose a framework to build generative classifier by density estimation, which considers the probability of the data into account.

  Despite much progress made on learning with label proportions, most of the previous work do not consider the knowledge transfer in the LLP problem. Transfer learning \cite{wang2011transfer,pan2010survey,tan2015transitive,Evgeniou2004Regularized} can transfer knowledge from the source task to the target task and the transferred knowledge can help target task to build a transfer learning classifier for prediction. In addition, transfer learning aims to solve new but similar problems effectively by utilizing previously acquired knowledge. Different from most of the previous work that considers the LLP problem as a single learning task in the training, we expect to build a transfer learning classifier for the target task by transferring knowledge from the source task. In addition, we may meet the uncertain data which is produced by sampling error or instrument imperfection \cite{aggarwal2009survey,zhou2018uncertain}. For example, in environmental monitoring applications, sensor networks typically create large amounts of uncertain data sets because of the noise in sensor inputs or errors in wireless transmission \cite{aggarwal2009survey}. Furthermore, some papers \cite{xiao2015robust,yang2016takagi,Deng2016Enhanced} that are related transfer learning and uncertainty modeling are proposed and show better performance. Hence, it is necessary to build a classifier on the target task for LLP problem by transferring knowledge from the source task where input data contain uncertain information.

  In this paper, we propose a novel approach to address the problem of transfer learning-based label proportions with uncertain data. In order to provide a more accurate classifier, this paper proposes a transfer learning-based approach for the problem of learning with label proportions(TL-LLP), which can transfer knowledge from source task to target task. In all, the main contributions of the paper can summarized as follows.
  \begin{itemize}
  \setlength{\itemsep}{0pt}
\setlength{\parsep}{0pt}
\setlength{\parskip}{0pt}
\item In the first step, we propose a transfer learning-based objective model by using a common parameter and incremental parameters to denote the direction of the classifiers for both tasks, which can transfer knowledge from the source task to the target task, and build a predictive classifier for learning with label proportions problem. Meanwhile, the proposed method can model the uncertain data using the reachability area, which indicates possible domain of the uncertain data. As a result, the proposed method can transfer knowledge for the LLP problem and reduce the effect of the uncertain data at the same time.
\item In the second step, in order to solve the proposed method, we propose an iterative framework to build the classifier and mitigate the effect of the uncertain data on the classifier. In addition, we present the update lemmas to refine the classifier and reduce the impact of the uncertain data. We further present the computation complex analysis of the proposed TL-LLP method. Thus, we can solve the problem of transfer learning-based label proportions where both tasks contain uncertain data.
\item We conduct extensive experiments on the data sets to investigate the performance of our proposed approach. The results have shown that our method performs better than existing LLP methods and is less sensitive to the uncertain data.
\end{itemize}

The remainder of this paper is structured as follows: in Section 2, we review the previous work related to our study. In the Section 3, the proposed method is introduced. Section 4 compares our method to the existing approaches. Finally, Section 5 concludes the work and presents the future work.
\section{Related work}
    In this section, we briefly review the previous work related to our study. We first introduce the methods for LLP problem in Section 2.1, and then review the previous methods on uncertain data in Section 2.2.
    \subsection{Learning with Label Proportions}
    To deal with LLP problem, many learning algorithms have been proposed. In the following, we briefly review the work on LLP problem based on support vector machine (SVM) model, probabilistic models and other models.

    Some methods are proposed to deal with LLP problem based on SVM technology. Rueping et al.\cite{rueping2010svm} treat the mean of each bag as a super-instance  and estimate a classifier based on support vector regression.  Yu et al.\cite{yu2013propto} introduce a large-margin framework called proportion-SVM which jointly optimizes over the unknown instances labels and the known label proportions. For the above framework, the two methods, called alter-$\propto$SVM, conv-$\propto$SVM, have been proposed for LLP problem. Following the idea in \cite{yu2013propto}, the method in \cite{wang2015linear} is based on twin SVM and needs to solve two smaller binary classification problems. Cui et al.\cite{cui2016laplacian} discuss how to combine the proportion learning framework with Laplacian term and analyze the structured information in proportion learning problem. The method introduces the Laplacian term to exploit the geometric information of data points. Chen et al.\cite{chen2017learning} try to address the LLP problem via nonparallel support vector machines, where the method can improve the classifiers to be a pair of nonparallel classification hyperplanes. Qi et al.\cite{qi2017learning} build a LLP-NPSVM method by a generalized classifier that determines instance labels according to two nonparallel hyper-planes under the supervision of label proportion information.

    Some methods are proposed to deal with  LLP problem based on probabilistic models. Hern{\'a}ndez et al.\cite{hernandez2011learning} adopt several versions of an  Expectation-Maximization algorithm to learn a naive Bayes model which assumes conditional independence between the predictive variables. Fan et al.\cite{fan2014learning} propose a new learning framework from the Bayesian perspective by estimating the conditional class density  to estimate the posterior probability. Meanwhile, with the deep belief networks model for estimating the log-probability, they rebuild the posterior probability for classification problem. Sun et al.\cite{sun2017probabilistic} build a probabilistic approach applied to the US presidential election, which uses cardinality potentials to perform exact inference over latent variables during learning, and introduces a novel message-passing algorithm to extend cardinality potentials to multivariate probability models.  Ardehaly et al.\cite{ardehaly2017mining} develop a models to estimate the relationship between political sentiment and demographics during the U.S. presidential election.

    In addition, Kuck et al.\cite{kuck2005learning} present a principled probabilistic model trained with an efficient Markov Chain Monte Carlo (MCMC) algorithm. Latter, MeanMap method which is based on modeling the conditional class probability has been proposed by Quadrianto et al.\cite{quadrianto2008estimating}. The MeanMap approach directly estimates the sufficient statistics of each bag by solving a linear system of equations. Stolpe et al.\cite{stolpe2011learning} introduce a method based on clustering with k-Means. Fish and Reyzin \cite{fish2017complexity} build an algorithm to solve foundational questions regarding the computational complexity of LLP, and also compare it with classical PAC learning to demonstrate the feasibility. Shi et al. \cite{shi2018learning} propose a algorithm called LLP-RF based on random forests, which has the advantage of dealing with high-dimensional LLP problem.

Despite the great progress made in this area, most existing work views the LLP problem as a single-task learning issue. However, in real-world applications, labelling a large amounts of data for new learning tasks may be expensive and time-consuming, and we expect to reduce the labeling efforts of the new task by transferring knowledge from related tasks. In this paper, we propose a novel approach TL-LLP, which not only processes data of uncertainty but also improves the performance of the new task’s classifier by transferring knowledge from related tasks.
    \subsection{Uncertain data}
    In the past, many learning approaches have been proposed to deal with data uncertainty. In the following, we briefly review previous work on uncertain data in clustering, classification and other application problems.

    Some  methods are devised to handle uncertain data in clustering problems. Kriegel and Pfeifle \cite{kriegel2005hierarchical} adopt the fuzzy distance functions to measure the problem of similarity between fuzzy objects and hierarchical density-based clustering algorithm. In the problem of clustering data objects whose locations are uncertain, Ngai et al.\cite{ngai2006efficient} propose UK-means algorithm to improve accuracy of the clusters formed in moving object uncertainty. Aggarwal and Charu \cite{aggarwal2007density} use multi-variate density estimation to handle error-prone and missing data. Xu et al.\cite{xu2015large} propose a clustering algorithm based on probability distribution similarity which aims at finding the largest margin between clusters. Zhou et al.\cite{zhou2018uncertain} study the problem of clustering distributed uncertain data  in distributed peer-to-peer network, in which the centralized global clustering solution is approximated by performing distributed clustering.

    To deal with uncertain data in classification problems, some methods are proposed as follow. Bi and Zhang \cite{bi2005support} adopt total support vector classification (TSVC) algorithm to deal with noisy input, which is motivated by the total least squares regression method. Gao and Wang \cite{gao2010direct} introduce a novel framework to help train either SVM or rule-based classifier. The algorithm mines discriminative patterns from uncertain data as classification features/rules. Tsang et al.\cite{tsang2011decision} restructure decision tree algorithms to handle data tuples with uncertain numerical attributes. Cao et al.\cite{cao2015classification} propose weighted ensemble classifier based on extreme learning machine algorithm, which can dynamically adjusts classifier and the weight to solve the problem of concept drift. Han et al.\cite{han2018two} discuss how to classify uncertain data streams, which handles both occurrence level and attribute level at the same time from positive and unlabeled examples.

    Recently, Denoeux and Thierry \cite{denoeux2013maximum} introduce a method  based on the maximization of a generalized likelihood criterion, which can be interpreted as a degree of agreement between the statistical model and the uncertain data. Liu et al.\cite{liu2013svdd} adopt a SVDD-based approach by introducing a confidence score for each input data point to detect outliers on uncertain data. Xiao et al.\cite{xiao2015robust} propose a method called uncertain one-class transfer learning, which is capable of constructing an accurate classifier on the target task by transferring knowledge from multiple source tasks whose data may contain uncertain observations. Islam et al.\cite{islam2018novel} describe how a belief-rule-based association rule is handled with the sensor data uncertainties.

    Most of the existing work on LLP problem does not consider the collected data may be corrupted with noises and contain uncertain information. In real-world applications, the data uncertainty produced by sampling error or instrument imperfection may reduce the performance of LLP problem. In this case, we will discuss how to construct an accurate classifier for LLP problem where the knowledge is transferred from the source task to target task, and both tasks may contain uncertain data.
\section{The Proposed Method}
    \subsection{Learning Setting}In learning with label proportions, a bag contains a set of instances, the label proportion of a bag is associated with the amount of positive instances in this bag. For convenience, we utilize capital letter $B_I$ and $P_I$ to denote the $Ith$ bag and its positive instances proportions. For the instances in a bag, we utilize lower letter $\bm{x}_i$ and $y_i$ to denote the $ith$ instance and the instance label.

    We denote the training sets of source task as $(B_I^{T_1},P_I^{T_1}), I=1,2,...t_1$, where $t_1$ is the number of bags for the source task $T_1$ and $P_I^{T_1}=|\{
\bm{x}_{1i} \in B_I^{T_1} : y_{1i}=1\}|/|B_I^{T_1}|$ is the estimate of the class probability $P(Y=1|B_I^{T_1})$. As a result, the training set of source task $D_1=\{\bm{x}_{11},\bm{x}_{12},\dots,\bm{x}_{1|D_1|}$\} is given in the form of $t_1$ disjoint bags:
    \begin{center}
     \mathsurround=0pt
    \setlength{\abovedisplayskip}{3pt}
    \setlength{\belowdisplayskip}{3pt}
    $ D_1=\{\bm{x}_{1i}|i \in B_I^{T_1}\}_{I=1}^{t_1},\quad B_m^{T_1}\cap B_n^{T_1} =\oslash ,\forall m\neq n$.\\
    \end{center}

    We focus on the binary case of $\mathcal{y}=\{1,-1\}$, if $P_I=0$, then all of the instances in $B_I$ are negative, if $P_I>0$, then the $B_I$ at least exists one positive instance. Let $(B_J^{T_2},P_J^{T_2}), J=1,2,...t_2$($D_2=\{\bm{x}_{21},\bm{x}_{22},\dots,\bm{x}_{2|D_2|}$\}) denote the training set for the target task $T_2$, and we have the same explanation. The similar operations as \cite{ruping2004simple,rueping2010svm}, we select a scaling function as bridge for bag proportion and label $y$ and use the Platt scaling function\cite{platt1999probabilistic} and inverse it as:

    \begin{equation}
     \mathsurround=0pt
    \setlength{\abovedisplayskip}{3pt}
    \setlength{\belowdisplayskip}{3pt}
    y=-log(\frac{1}{p}-1)
    \end{equation}
    For estimate a linear classification function $f(\bm{x})=\bm{w}^T\bm{x}+b$ well, we  require that $f$ predicts $y$ on average:
    \begin{equation}
     \mathsurround=0pt
    \setlength{\abovedisplayskip}{3pt}
    \setlength{\belowdisplayskip}{3pt}
    \forall_i: \frac{1}{|B_i|}\sum_{j\in B_i}(\bm{w^Tx_j}+b)\approx y_i
    \end{equation}
\begin{figure}[t]
    \centering
    \includegraphics[scale=0.35]{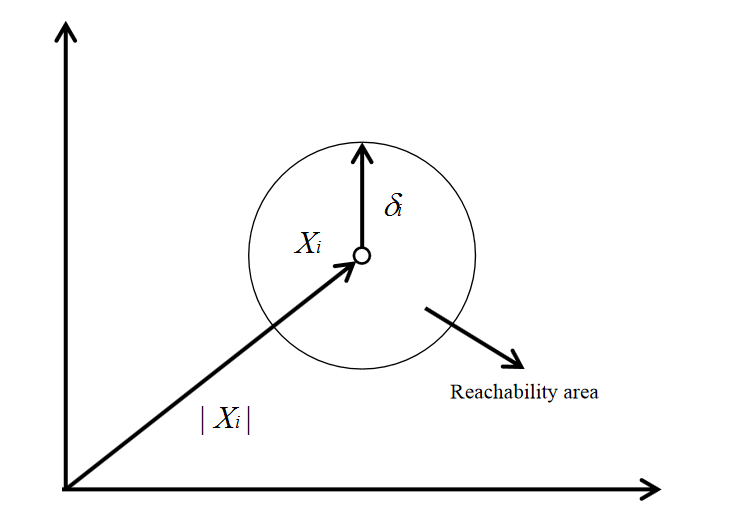}
    \caption{Illustration of reachability area of instance $X_i$}
    \label{fig:area}
\end{figure}
\subsection{ The proportion transfer learning framework}
In this section, we will introduce our proposed method. For two related tasks, we construct an SVR-based classifier to transfer knowledge from the source task to the target task. We assume that the models of two tasks are all close to a similar model, and then we train SVR on $(B_I^{T_1},P_I^{T_1})$ for the source task and on $(B_J^{T_2},P_J^{T_2})$ for the target task. In order to transfer knowledge from the source task to the target task, we make
\begin{equation}
 \mathsurround=0pt
\setlength{\abovedisplayskip}{1pt}
\setlength{\belowdisplayskip}{1pt}
   \bm{w}_1=\bm{w}_0+\bm{v}_1 \quad and \quad \bm{w}_2=\bm{w}_0+\bm{v}_2
\end{equation}
where $\bm{w}_0$ is a parameter to facilitate the transfer and the specific parameter $\bm{v}_1$, $\bm{v}_2$ represent the discrepancy between the local optimal decision boundary and the globe optimal decision boundary($\bm{w}_0$). Let $f_1(\bm{x})={\bm{w}_1}^{T}\bm{x}+b_1$ and $f_2(\bm{x})={\bm{w}_2}^{T}\bm{x}+b_2$ be the classification planes for source task $T_1$ and target task $T_2$. For the consideration of data uncertainty, we assume each input data $\bm{x}_i$ is subject to an additive noise vector $\triangle\bm{x}_i$. As studied in \cite{Huffel1991The,bi2004support,liu2014uncertain}, we consider a simple bound score for each sample such that:
  \begin{equation}　
\|\triangle\bm{x}_i\| \leq \delta_i.
  \end{equation}
This setting has a similar influence of assuming $\triangle \bm{x}_i$ has a certain distribution. For example, we assume that $\triangle \bm{x}_i$ follows a Gaussian noise model:
 \begin{equation*}
 \mathsurround=0pt
\setlength{\abovedisplayskip}{1pt}
\setlength{\belowdisplayskip}{1pt}
   p(|\bm{x}_i-\bm{x}_i^s|)\sim exp(\frac{||\bm{x}_i-\bm{x}_i^s||^2}{2\sigma^2}),
\end{equation*}
The bound $\delta_i$ has an effect similar to the standard deviation $\sigma$ in the Gaussian noise model. In addition, the constraint $||\triangle\bm{x}_i|| \leq \delta_i$ replaces the squared penalty term $\frac{||\bm{x}_i-\bm{x}_i^s||^2}{2\sigma^2}$.

 We let $\bm{x}_i+\triangle\bm{x}_i$ denote the reachability area of example $\bm{x}_i$, as illustrated in Figure \ref{fig:area}. Thus, the original uncorrupted input $\bm{x}_i^s$ can thereafter be denoted as $\bm{x}_i^s=\bm{x}_i+\triangle\bm{x}_i$, and $\bm{x}_i^s$ falls into the reachability area of $\bm{x}_i$. The metric of this uncertain model is if we can obtain $\triangle \bm{x}_i$, the uncorrupted data can be estimated. We then consider two similar tasks learning to solve the LLP problem as follows:

\begin{equation}\label{equ:problem}
\setlength{\abovedisplayskip}{3pt}
\setlength{\belowdisplayskip}{3pt}
    \begin{split}
    \min\quad &\frac{1}{2}||\bm{w}_0||{^2}+\frac{\lambda_1}{2}||\bm{v}_1||{^2}+\frac{\lambda_2}{2}||\bm{v}_2||{^2}\\
    &+C_1\sum\limits_{i=1}^{t_1}(\xi_{1i}+\xi_{1i}^*)+C_2\sum\limits_{m=1}^{t_2}(\xi_{2m}+\xi_{2m}^*)
    \end{split}
  \end{equation}
Subject to:
\begin{equation*}
\begin{split}
\forall_{i=1}^{t_1}:&
            \begin{array}{lcl}
             {\frac{1}{|B_i^{T_1}|}\sum\limits_{j\in B_i^{T_1} }(\bm{w}_1^{T}(\bm{x}_{1j}+\triangle\bm{x}_{1j})+b_1)-y_{1i} \leq \varepsilon_{1i} + \xi_{1i}} \\
             {y_{1i} -\frac{1}{|B_i^{T_1}|}\sum\limits_{j\in B_i^{T_1}}(\bm{w}_1^{T}(\bm{x}_{1j}+\triangle\bm{x}_{1j})+b_1) \leq \varepsilon_{1i} + \xi_{1i}^*}
             \end{array}
\end{split}
\end {equation*}
\begin{equation*}
\begin{split}
\forall_{m=1}^{t_2}:&
\begin{array}{lcl}
             {\frac{1}{|B_m^{T_2}|}\sum\limits_{n\in B_m^{T_2} }(\bm{w}_2^{T}(\bm{x}_{2n}+\triangle\bm{x}_{2n})+b_2)-y_{2m} \leq \varepsilon_{2m} + \xi_{2m}} \\
              {  y_{2m} -\frac{1}{|B_m^{T_2}|}\sum\limits_{n\in B_m^{T_2}}(\bm{w}_2^{T}(\bm{x}_{2n}+\triangle\bm{x}_{2n})+b_2) \leq \varepsilon_{2m} + \xi_{2m}^*}
             \end{array}\\
\end{split}
\end {equation*}
\begin{equation*}
\triangle\bm{x}_{1j}\leq \delta_{1j},\quad \triangle\bm{x}_{2n}\leq \delta_{2n},\quad \xi_{1i} ,\xi_{1i}^*\geq 0,\quad \xi_{2m},\xi_{2m}^*  \geq 0
\end {equation*}
\begin{equation*}
i=1,\dots ,t_1,\quad j=1,\dots ,|D_1|,\quad m=1,\dots ,t_2,\quad n=1,\dots ,|D_2|
\end {equation*}

where $\xi_{ti}$ and $\xi_{ti}^*(t=1,2)$ are training errors; parameter $\lambda_1,\lambda_2>0$ control the tradeoff between the source task and target task; $C_1$ and $C_2$ are parameter to balance the margin and training errors; the $\varepsilon_{1i}$ and $\varepsilon_{2i}$ controls the size of $\varepsilon$-insensitive zone. Since we can determine a choice of $\triangle\overline{\bm{x}}_i$ to render $\bm{x}_i+\triangle\bm{x}_i$, the method of modeling noise lets TL-LLP less sensitive to the sample corrupted by noise.

To solve the problem (\ref{equ:problem}), we use an iterative approach to calculate $\bm{w}_0,\bm{v}_t,b_t,\triangle\bm{x}_{ti},\xi_{ti}$ and $\xi_{ti}^*(t=1,2)$ to obtain the classifier. In the first step, we fix each $\triangle\bm{x}_{1i}$ and $\triangle\bm{x}_{2i}$, and solve (\ref{equ:problem}) to obtain $\bm{w}_0, \bm{v}_t, b_t, \xi_{ti}$ and $\xi_{ti}^*(t=1,2)$; in the second step, we fix $\bm{w}_0, \bm{v}_t, b_t, \xi_{ti}$ and $\xi_{ti}^*(t=1,2)$ to calculate the values of $\triangle\bm{x}_{1i}$ and $\triangle\bm{x}_{2i}$. In the following, we detail the  above two steps and the detailed derivations refer to Appendix section.

First, we fix each $\triangle\bm{x}_{1i}$ and $\triangle\bm{x}_{2j}$ as $\triangle\overline{\bm{x}}_{1i}$ and $\triangle\overline{\bm{x}}_{2j}$, respectively, and let them to a small value such that $||\triangle\overline{\bm{x}}_{1i}||\leq\delta_{1i}$ and $||\triangle\overline{\bm{x}}_{2j}||\leq\delta_{2j}$, then we solve the optimization problem (\ref{equ:problem}) and get the following lemma.
\newtheorem{lemma}{Lemma}
\begin{lemma} \label{lemma1}If initialize each $\triangle\overline{\bm{x}}_{1i}$ and $\triangle\overline{\bm{x}}_{2j}$ as zero vector to satisfy the constraints, the solution of problem (\ref{equ:problem}) is transformed into
\begin{equation}\label{equ:prime}
\setlength{\abovedisplayskip}{3pt}
\setlength{\belowdisplayskip}{3pt}
    \begin{split}
    min\quad &\frac{1}{2}||\bm{w}_0||{^2}+\frac{\lambda_1}{2}||\bm{v}_1||{^2}+\frac{\lambda_2}{2}||\bm{v}_2||{^2}\\
    &+C_1\sum\limits_{i=1}^{t_1}(\xi_{1i}+\xi_{1i}^*)+C_2\sum\limits_{m=1}^{t_2}(\xi_{2m}+\xi_{2m}^*)
    \end{split}
  \end{equation}
Subject to:
\begin{equation*}
\begin{split}
\forall_{i=1}^{t_1}:&
            \begin{array}{lcl}
             {\frac{1}{|B_i^{T_1}|}\sum\limits_{j\in B_i^{T_1} }(\bm{w}_1^{T}(\bm{x}_{1j}+\triangle\overline{\bm{x}}_{1j})+b_1)-y_{1i} \leq \varepsilon_{1i} + \xi_{1i}} \\
             {y_{1i} -\frac{1}{|B_i^{T_1}|}\sum\limits_{j\in B_i^{T_1}}(\bm{w}_1^{T}(\bm{x}_{1j}+\triangle\overline{\bm{x}}_{1j})+b_1) \leq \varepsilon_{1i} + \xi_{1i}^*}
             \end{array}
\end{split}
\end {equation*}
\begin{equation*}
\begin{split}
\forall_{m=1}^{t_2}:&
\begin{array}{lcl}
             {\frac{1}{|B_m^{T_2}|}\sum\limits_{n\in B_m^{T_2} }(\bm{w}_2^{T}(\bm{x}_{2n}+\triangle\overline{\bm{x}}_{2n})+b_2)-y_{2m} \leq \varepsilon_{2m} + \xi_{2m}} \\
              {  y_{2m} -\frac{1}{|B_m^{T_2}|}\sum\limits_{n\in B_m^{T_2}}(\bm{w}_2^{T}(\bm{x}_{2n}+\triangle\overline{\bm{x}}_{2n})+b_2) \leq \varepsilon_{2m} + \xi_{2m}^*}
             \end{array}\\
\end{split}
\end {equation*}
\begin{equation*}
\xi_{1i},\xi_{1i}^*  \geq 0,\quad\xi_{2m} ,\xi_{2m}^*\geq 0,\quad i=1,\dots ,t_1, \quad m=1,\dots ,t_2.
\end {equation*}
\end{lemma}

Since the constraint is released and the problem (\ref{equ:prime}) is a QP problem,  which can be solved via the dual form. Second, we need to resolve optimization problem (\ref{equ:prime}), and have Lemma 2.
\begin{lemma}\label{lemma2}
By introducing the Lagrangian function \cite{pednault1997statistical}, the solution of the optimization problem (\ref{equ:prime}) can be given by:
\begin{align}\label{equ:dual}
    \begin{split}
    min\quad
    &\frac{1+\lambda_1}{2\lambda_1}\sum\limits_{i,j=1}^{t_1}\frac{(\alpha_{1i}^*-\alpha_{1i})(\alpha_{1j}^*-\alpha_{1j})}{|B_i^{T_1}||B_j^{T_1}|}\sum\limits_{i^\prime\in B_i^{T_1},j^\prime\in B_j^{T_1}}K(\overline{\bm{x}}_{1i^\prime},\overline{\bm{x}}_{1j^\prime})\\
    &+\frac{1+\lambda_2}{2\lambda_2}\sum\limits_{m,n=1}^{t_2}\frac{(\alpha_{2m}^*-\alpha_{2m})(\alpha_{2n}^*-\alpha_{2n})}{|B_m^{T_2}||B_n^{T_2}|}\sum\limits_{m^\prime\in B_m^{T_2},n^\prime\in B_n^{T_2}}K(\overline{\bm{x}}_{2m^\prime},\overline{\bm{x}}_{2n^\prime})\\
    &+\sum\limits_{i=1}^{t_1}\sum\limits_{m=1}^{t_2}\frac{(\alpha_{1i}^*-\alpha_{1i})(\alpha_{2m}^*-\alpha_{2m})}{|B_i^{T_1}||B_m^{T_2}|}\sum\limits_{i^\prime\in B_i^{T_1},m^\prime\in B_m^{T_2}}K(\overline{\bm{x}}_{1i^\prime},\overline{\bm{x}}_{2m^\prime})\\
    &-\sum\limits_{i=1}^{t_1}(y_{1i}(\alpha_{1i}^*-\alpha_{1i})-\varepsilon_{1i}(\alpha_{1i}^*+\alpha_{1i}))-\sum\limits_{m=1}^{t_2}(y_{2m}(\alpha_{2m}^*-\alpha_{2m})-\varepsilon_{2m}(\alpha_{2m}^*+\alpha_{2m}))
    \end{split}
\end{align}
Subject to:\\
\begin{equation*}
\sum\limits_{i=1}^{t_1}(\alpha_{1i}-\alpha_{1i}^*)+\sum\limits_{m=1}^{t_2}(\alpha_{2m}-\alpha_{2m}^*)=0
\end{equation*}
\begin{equation*}
\forall_{i=1}^{t_1}:0 \leq \alpha_{1i}, \alpha_{1i}^{*} \leq C_1
\end{equation*}
\begin{equation*}
\forall_{m=1}^{t_2}:0 \leq \alpha_{2m}, \alpha_{2m}^{*} \leq C_2
\end{equation*}
then we can obtain the solutions of Lagrange multipliers and the values of  $\overline{\bm{w}}_0$, $\overline{\bm{v}}_1$ and $\overline{\bm{v}}_2$ can be calculated as
\begin{align}
&\overline{\bm{w}}_0=\sum\limits_{i=1}^{t_1}\frac{(\alpha_{1i}^*-\alpha_{1i})}{|B_i^{T_1}|}\sum\limits_{j\in B_i^{T_1}}\overline{\bm{x}}_{1j}+\sum\limits_{m=1}^{t_2}\frac{(\alpha_{2m}^*-\alpha_{2m})}{|B_m^{T_2}|}\sum\limits_{n\in B_m^t}\overline{\bm{x}}_{2n}\\
&\overline{\bm{v}}_1=\frac{1}{\lambda_1}\sum\limits_{i=1}^{t_1}(\alpha_{1i}^*-\alpha_{1i})\frac{1}{|B_i^{T_1}|}\sum\limits_{j\in B_i^{T_1} }\overline{\bm{x}}_{1j}\\
&\overline{\bm{v}}_2=\frac{1}{\lambda_2}\sum\limits_{m=1}^{t_2}(\alpha_{2m}^*-\alpha_{2m})\frac{1}{|B_m^{T_2}|}\sum\limits_{n\in B_m^{T_2}}\overline{\bm{x}}_{2n}
\end{align}
where $\alpha_{1i},\alpha_{1i}^{*}$,$\alpha_{1m}$ and $\alpha_{1m}^{*}$ are Lagrange multipliers; it has $\overline{\bm{x}}_{1i}=\bm{x}_{1i}+\triangle\overline{\bm{x}}_{1i}$; $\overline{\bm{x}}_{1j},\overline{\bm{x}}_{2m}$ and $\overline{\bm{x}}_{2n}$ are similar to $\overline{\bm{x}}_{1i}$.
\end{lemma}
After fix $\triangle\overline{\bm{x}}_{1i}$ and $\triangle\overline{\bm{x}}_{2m}$ to obtain $\overline{f}_1(\bm{x})$ and $\overline{f}_2(\bm{x})$, the next step is to fix $\overline{f}_1(\bm{x})$ and $\overline{f}_2(\bm{x})$ to calculate new $\triangle\overline{\bm{x}}_{1i}$ and $\triangle\overline{\bm{x}}_{2m}$. We have Lemma 3 as follows.
\begin{lemma} \label{lemma3}
By fixing two hyperplanes $\overline{f}_1(\bm{x})$ and $\overline{f}_2(\bm{x})$, the solutions of $\triangle\overline{\bm{x}}_{ti}$(t=1,2) for optimizing problem (\ref{equ:problem}) are

\begin{equation}\label{equ:tria}
\triangle\overline{\bm{x}}_{ti} =
\begin{cases}
\delta_{ti}\frac{-\bm{u}_t}{||\bm{u}_t||} & \text{if } \overline{f}_t(\bm{x}_{ti})-y_{ti}>\varepsilon,\\
0 & \text{if } |\overline{f}_t(\bm{x}_{ti})-y_{ti}|<\varepsilon,\\
\delta_{ti}\frac{\bm{u}_t}{||\bm{u}_t||} & \text{if } y_{ti}-\overline{f}_t(\bm{x}_{ti})>\varepsilon.
\end{cases}
 i = 1,...,|D_t|, t = 1,2.
\end{equation}

where it has
\begin{equation*} \bm{u}_1=\frac{1+\lambda_1}{\lambda_1}\sum\limits_{j=1}^{t_1}\frac{(\alpha_{1j}^*-\alpha_{1j})}{|B_j^s|}\sum\limits_{k\in B_j^s}K^{\prime}(\bm{x}_{jk}+\triangle\bm{x}_{jk},\bm{x}_i)+\sum\limits_{m=1}^{t_2}\frac{(\alpha_{2m}^*-\alpha_{2m})}{|B_m^t|}\sum\limits_{n\in B_m^t}K^{\prime}(\bm{x}_{mn}+\triangle\bm{x}_{mn},\bm{x}_i)
 \end{equation*}
\begin{equation*}
\bm{u}_2=\sum\limits_{j=1}^{t_1}\frac{(\alpha_{1j}^*-\alpha_{1j})}{|B_j^{T_1}|}\sum\limits_{k\in B_j^{T_1}}K^{\prime}(\bm{x}_{1k}+\triangle\bm{x}_{1k},\bm{x}_i)+\frac{1+\lambda_2}{\lambda_2}\sum\limits_{m=1}^{t_2}\frac{(\alpha_{2m}^*-\alpha_{2m})}{|B_m^{T_2}|}\sum\limits_{n\in B_m^{T_2}}K^{\prime}(\bm{x}_{2n}+\triangle\bm{x}_{1n},\bm{x}_i).
 \end{equation*}

The theorem indicates that, for given $\overline{f}_1(\bm{x})$ and $\overline{f}_2(\bm{x})$, the minimization of problem (\ref{equ:problem}) over $\triangle\overline{\bm{x}}$ is quite straightforward.
\end{lemma}

\begin{algorithm}[htb]
\caption{TL-LLP}
\label{alg:Framwork}
\begin{algorithmic}[1] 
\REQUIRE ~~\\ 
The source dataset and target dataset: $(B_I^s,P_I^s)$ and $(B_ Ｊ^t,P_J^t)$;\\
Parameter of TL-LLP: $C_1,C_2,\lambda_1,\lambda_2,\varepsilon$;\\
Bound value for each sample: $\delta_i$;
\ENSURE  $f_1(\bm{x})$ and $f_2(\bm{x})$;
\STATE Initialize each $\triangle\overline{\bm{x}}_{1i}$=0 and $\triangle\overline{\bm{x}}_{2m}$=0;
\STATE $t=0$;
\STATE Initialize $F_{val}(t)=\propto$;
\REPEAT
\STATE $t=t+1$;
\STATE Fix $\triangle\overline{\bm{x}}_{1i}$ and $\triangle\overline{\bm{x}}_{2m}$ and solve problem (\ref{equ:prime});
\STATE Let $F(t)=F(\alpha)$;
\STATE Compute $\overline{f}_1(\bm{x})$ and $\overline{f}_2(\bm{x})$ based on Equations (\ref{equ:w})-(\ref{equ:v2});
\STATE Fix $\overline{f}_1(\bm{x})$ and $\overline{f}_2(\bm{x})$ and resolve optimization problem (\ref{equ:problem}) to update each $\triangle\overline{\bm{x}}_{1i}$ $\triangle\overline{\bm{x}}_{2m}$  according to (\ref{equ:tria});
\STATE Let $F_{max}=max\{|F_{val}(t-1)|,|F_{val}(t)|\}$;
\UNTIL{$|F_{val}(t)-F_{val}(t-1)|<\epsilon|F_{max}|$}
\RETURN $f_1(\bm{x})=\bm{w}_2^T\bm{x}+b_1$ and $f_2(\bm{x})=\bm{w}_2^T\bm{x}+b_2$; 
\end{algorithmic}
\end{algorithm}
After that, we have one round of alternation and continue to update $\overline{f}_1(\bm{x})$, $\overline{f}_2(\bm{x})$ and $\triangle\overline{\bm{x}}$ iteratively. By referring to the alternating optimization method in \cite{Huffel1991The}, we propose an iterative framework to resolve problem (\ref{equ:problem}) in Algorithm 1. By employing the stopping criterion as in \cite{wang2010linear}, since the values of $F_{val}(t)$ is nonnegative, the algorithm will be stopped when $|F_{val}(t)-F_{val}(t-1)|/|F_{max}|$ is smaller than the threshold $\epsilon$.

  For the computation complex of the proposed TL-LLP method, assume that training a standard SVM method requires $O([$number of training data$]^2 )$ time. In the paper, solving the optimization problem (\ref{equ:dual}) requires solving a standard SVM problem with $|D_1|$ source task data and $|D_2|$ task target data. The update of $\triangle\overline{\bm{x}}$ in (\ref{equ:tria}) just needs linear time, that is $O(|D_1|+|D_2|)$. Suppose the iterative approach stops after $m$ times iterations. Thus, the overall complexity in solving the problem (\ref{equ:dual}) is $m\cdot O((|D_1|+|D_2|)^2)+m\cdot O(|D_1|+|D_2|)$. For the prediction of the transfer learning classifier, after solving the problem (\ref{equ:problem}), we can obtain the transfer classifier $f_2(\bm{x})=(\bm{w}_0+\bm{v}_2)^T\cdot \bm{x}+b_2$. For the instance $\bm{x}_{1i}$ in target task, if $f_1(\bm{x}_{1i})=\bm{w}_2^T\cdot \bm{x}_{1i}+b_2>0$, the instance is labeled as positive; otherwise, it's labeled as negative.
  
  \begin{table}[tbp]
     \centering
     \resizebox{\textwidth}{25mm}{
    \begin{tabular}{lllcllc}
    \hline
    Dataset ID&Source Task&Size&Attributes&Target task&Size&Attributes\\
    \hline
    Dataset 1&Com-wind.misc&2000&200&Com-wind.x&800&200\\
    Dataset 2&Com.pc.hardware&2000&200&Com.mac.hardware&800&200\\
    Dataset 3&Sci-elec&2000&200&Sci-med&800&200\\
    Dataset 4&Rec.sport.baseball&2000&200&Rec.sport.hockey&800&200\\
    Dataset 5&Rec.autos&2000&200&Rec.motorcycles&800&200\\
	Dataset 6&Talk.politics.misc&2000&200&Talk.politics.guns&800&200\\
    Dataset 7&People(1)&1800&240&People(2)&600&240\\
    Dataset 8&Orgs(1)&1800&240&Orgs(2)&600&240\\
    Dataset 9&People(1)&1800&240&Places(1)&600&240\\
    \hline
    \end{tabular}}
    \caption{Datasets used in the Experiments}
    \end{table}
\section{Experiments}
In this section, we will conduct experiments to investigate the performance of the proposed TL-LLP method. These experiments were done on a laptop with 2.4 GHz CPU and 4GB RAM.
\subsection{Baselines and Metrics}
 For comparison, another four state-of-the-art LLP methods, MeanMap \cite{quadrianto2009estimating}, Inverse Calibration(InvCal) \cite{rueping2010svm}, alter proportion-SVM(alter-$\propto$SVM) \cite{yu2013propto} and convex proportion-SVM(conv-$\propto$SVM) \cite{yu2013propto} are used as baselines.
\begin{itemize}
\setlength{\itemsep}{0pt}
\setlength{\parsep}{0pt}
\setlength{\parskip}{0pt}
\item MeanMap \cite{quadrianto2009estimating}:  the first method estimates the mean of each class using the mean of each bag and the label proportions, and employs an Markov Chain Monte Carlo algorithm to handle the problem.
\item InvCal \cite{rueping2010svm}: the second method replaces the whole bag with mean of each bag and trains the classifier by combining support vector regression and inverse classifier calibration.
\item Alter-$\propto$SVM \cite{yu2013propto}: the third method is an alternating optimization method for the non-convex integer programming problem($\propto$SVM). This baseline trains a classifier by iteratively optimizing (w, b) and y until the objective converges.
\item  Conv-$\propto$SVM \cite{yu2013propto}: the fourth method is a convex relaxation method, which transforms the $\propto$SVM formulation to a convex function and does not require multiple initializations.
\end{itemize}

 In general, the problem of learning with label proportions always use accuracy to evaluate the performance of the methods, we use accuracy as a measure of metrics in the experiments.

 \subsection{Data Sets and Settings}
To evaluate the properties of our method, we compare the performance of different baselines on 20 Newsgroups\footnote{\url{http://people.csail.mit.edu/jrennie/20Newsgroups/.}} and Reuters-21578\footnote{\url{http://www.daviddlewis.com/resources/testcollections/.}}, which are popularly used in the previous transfer learning work \cite{dai2007transferring,zhong2009cross,pan2010survey}. In 20 Newsgroups corpus, there are 20 sub-categories under each top category and each sub-category has 1000 samples. Similarly, Reuters-21578 contains Reuters news wire articles which are split into five top categories where each category includes different sub-categories. For example, the top category ``place'' has 175 sub-categories.
  
\begin{figure}[tbp]
\centering
\begin{minipage}[t]{0.49\textwidth}
\centering
\includegraphics[width=8.5cm]{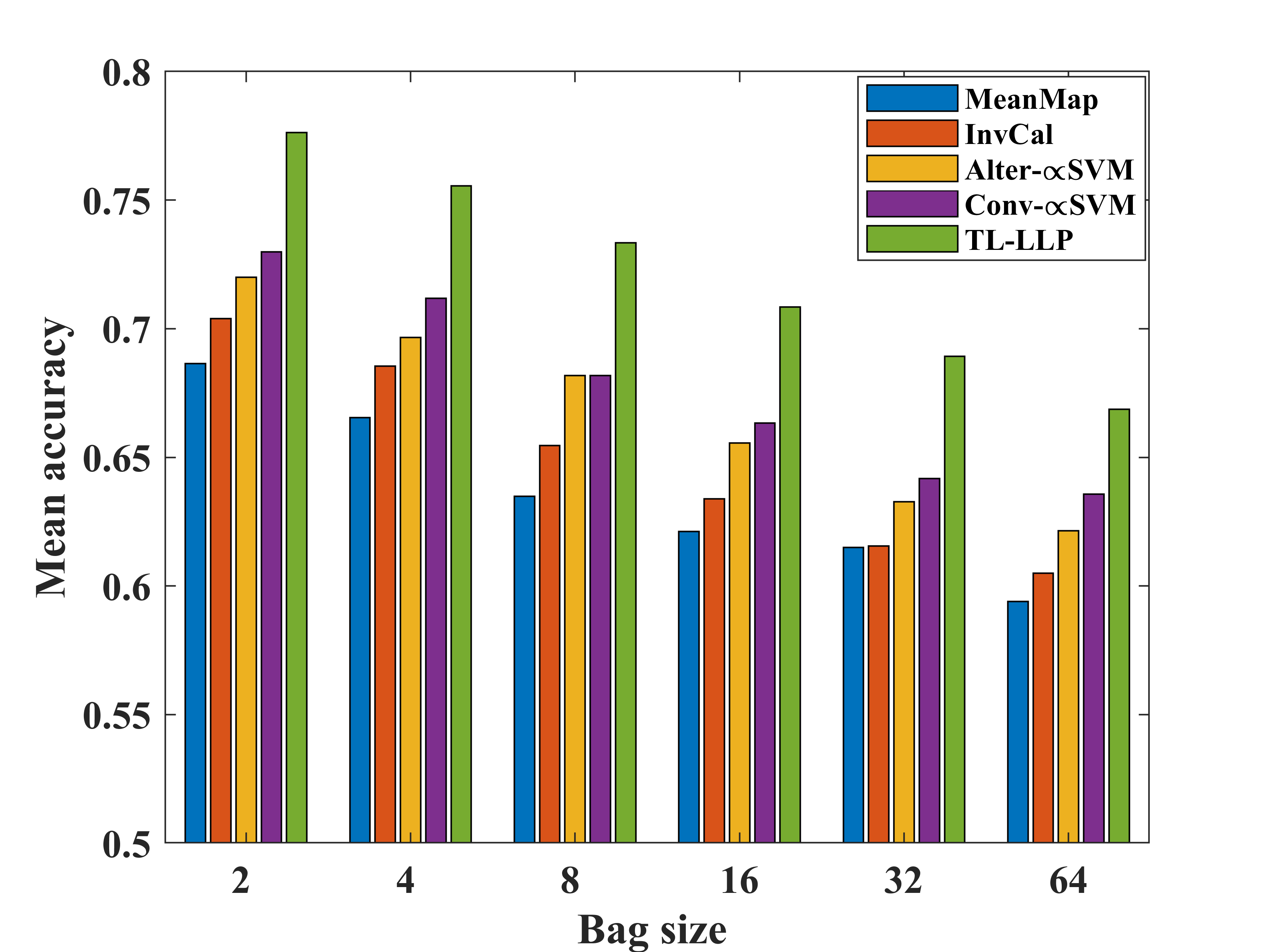}
\caption{The mean accuracy of the data sets}
\label{fig:meac}
\end{minipage}
\begin{minipage}[t]{0.49\textwidth}
\centering
\includegraphics[width=8.5cm]{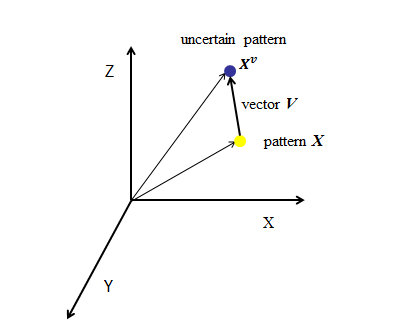}
\caption{Illustration of adding noises to the data example: {\bfseries x} is the original data example, {\bfseries v} is a noise vector, $\bf{x^v}$ is the new data example with added noises.}
\label{fig:noise}
\end{minipage}
\end{figure}

Since the two datasets are not designed for LLP problem, similar to the operations in\cite{zhang2008m3miml,zhou2009multi,liu2018selective}, we reorganize the LLP datasets based on the top categories. First, we choose a sub-category $\alpha$(1) from a top category (A) as a positive class, so each sample in this sub-category $\alpha$(1) is seen as a positive sample and other sub-categories are negative class. Second, for the source task, we randomly select a number of samples from the positive sub-category $\alpha$(1) as positive instances and the same number samples from other sub-categories as negative instances, and form them as a source task. The same operation is conducted on the target task to form the positive instances and the negative instances.  In order to make the two tasks relevant, we let the positive class of the source task and the target task have the same top category in the generation, such as $\alpha$(1) for the source task, $\alpha$(2) for the target task. Without loss of effectiveness, we only retain the words with higher document frequency to reduce dimensionality and each instance is represented by features.

Using the above operations, we generate nine data sets for the source task and target task and listed both tasks in Table 2. In Table 2, both data sets of source task and target task will be experimented by the proposed method, and the other methods will only conduct experiment using the target task since they are proposed for single task learning.

For the experiment setting, we conduct similar to the previous LLP work \cite{rueping2010svm,yu2013propto}, we randomly select the instances into bags of a fixed size and choose the bag sizes of 2, 4, 8, 16, 32, 64. In our approach, $\lambda_1$ and $\lambda_2$ control the trade-off between source task and target task. Since we are more concerned with the target task than the source task in transfer learning, we set $\lambda_1 > \lambda_2$, $\varepsilon \in[0,1]$, $C_i\in[2^{-2},2^7]$(i=1,2) and the bond score $\delta_i$ is set 0.01. For other baselines, we set the experiments similar to their own works. For MeanMap, let the parameter $\lambda \in\{0.1,1,10\}$. For InvCal, the parameters are tuned from $C\in[2^{-2},2^5]$, and $\varepsilon \in \{0.01,0.1\}$. For alter-$\propto$SVM, the parameters are tuned from $C \in [2^{-2},2^5]$, and $C_p \in[2^{-2},2^7]$. For conv-$\propto$SVM, the parameters are tuned from $C\in[2^{-2},2^5]$, and $\varepsilon \in \{0.01,0.1\}$.

We utilize linear kernels $k(x_1,x_2)=x_1\cdot x_2$ since it always performs well for text classification\cite{sebastiani2002machine}. Then, we execute experiments with 5-fold cross validation, and the performance is evaluated by accuracy. We then report the mean accuracies with standard deviations of the five times of testing.
\begin{table}[!ht]
\caption{Accuracy with linear kernel obtained by MeanMap, InvCal, Alter-$\propto$SVM, Conv-$\propto$SVM and TL-LLP methods.}
\label{tab:acc}
\newcommand{\tabincell}[2]{\begin{tabular}{@{}#1@{}}#2\end{tabular}}
\resizebox{\textwidth}{75mm}{
    \begin{tabular}{p{10pt}|ccccccc}
    \hline
    ID&Method&2&4&8&16&32&64\\
    \hline
    1&\tabincell{c}{MeanMap \\ InvCal\\Alter-$\propto$SVM\\Conv-$\propto$SVM\\ TL-LLP} &\tabincell{c}{65.41$\pm$1.28\\
66.08$\pm$0.24\\
70.51$\pm$1.02\\
75.22$\pm$0.33\\
\bf{78.91$\pm$0.22}
}
&\tabincell{c}{63.14$\pm$0.99\\
64.95$\pm$0.75\\
69.33$\pm$1.11\\
73.57$\pm$1.05\\
\bf{77.35$\pm$0.63}
}
&\tabincell{c}{56.36$\pm$0.69\\
60.17$\pm$1.87\\
65.62$\pm$1.39\\
71.33$\pm$1.44\\
\bf{75.02$\pm$0.56}
} &\tabincell{c}{55.44$\pm$1.91\\
58.82$\pm$1.46\\
63.78$\pm$2.58\\
69.74$\pm$1.02\\
\bf{73.25$\pm$1.07}
} &\tabincell{c}{55.08$\pm$1.37\\
58.62$\pm$1.28\\
61.52$\pm$3.20\\
70.48$\pm$1.38\\
\bf{71.09$\pm$1.03}
} &\tabincell{c}{55.25$\pm$1.72\\
58.27$\pm$1.44\\
60.27$\pm$3.55\\
65.86$\pm$2.68\\
\bf{67.02$\pm$1.34}
}\\\hline
2&\tabincell{c}{MeanMap \\ InvCal\\Alter-$\propto$SVM\\ Conv-$\propto$SVM\\ TL-LLP}
 &\tabincell{c}{66.42$\pm$0.88\\
69.73$\pm$2.48\\
68.34$\pm$1.23\\
71.43$\pm$0.53\\
\bf{75.68$\pm$0.62}
} &\tabincell{c}{64.86$\pm$0.38\\
66.53$\pm$1.33\\
65.35$\pm$0.35\\
68.83$\pm$0.80\\
\bf{73.94$\pm$0.26}\\
}  &\tabincell{c}{61.89$\pm$0.43\\
63.19$\pm$0.68\\
64.01$\pm$0.37\\
65.69$\pm$0.45\\
\bf{71.04$\pm$0.31}
} &\tabincell{c}{61.37$\pm$1.16\\
61.20$\pm$0.49\\
60.32$\pm$0.22\\
63.62$\pm$1.52\\
\bf{68.96$\pm$0.18}
} &\tabincell{c}{60.75$\pm$1.22\\
59.32$\pm$0.27\\
59.52$\pm$0.37\\
62.65$\pm$1.50\\
\bf{66.97$\pm$0.24}
}&\tabincell{c}{59.78$\pm$0.53\\
59.48$\pm$1.21\\
60.07$\pm$0.42\\
62.18$\pm$1.22\\
\bf{64.40$\pm$0.43}
}\\\hline
    3&\tabincell{c}{MeanMap \\ InvCal\\Alter-$\propto$SVM\\ Conv-$\propto$SVM\\ TL-LLP} &\tabincell{c}{62.33$\pm$0.70\\
65.37$\pm$2.48\\
63.34$\pm$1.23\\
68.43$\pm$1.33\\
\bf{73.01$\pm$1.02}
} &\tabincell{c}{60.62$\pm$0.61\\
64.59$\pm$3.47\\
61.59$\pm$2.85\\
67.05$\pm$0.83\\
\bf{71.05$\pm$0.83}
} &\tabincell{c}{59.02$\pm$1.22\\
62.51$\pm$1.53\\
60.01$\pm$3.47\\
64.02$\pm$0.78\\
\bf{69.02$\pm$0.68}
} &\tabincell{c}{57.76$\pm$1.73\\
60.07$\pm$2.56\\
59.23$\pm$4.73\\
63.28$\pm$1.52\\
\bf{67.54$\pm$1.62}
} &\tabincell{c}{56.37$\pm$1.35\\
58.62$\pm$3.02\\
59.52$\pm$4.22\\
60.68$\pm$1.50\\
\bf{66.42$\pm$1.02}
} &\tabincell{c}{56.10$\pm$2.03\\
58.21$\pm$4.01\\
57.72$\pm$4.78\\
60.28$\pm$1.52\\
\bf{65.46$\pm$1.33}
}\\\hline
    4&\tabincell{c}{MeanMap \\ InvCal\\Alter-$\propto$SVM\\ Conv-$\propto$SVM\\ TL-LLP}
    &\tabincell{c}{78.25$\pm$0.81\\
78.18$\pm$0.12\\
81.45$\pm$1.23\\
82.32$\pm$0.65\\
\bf{86.21$\pm$0.34}
} &\tabincell{c}{75.47$\pm$0.72\\
75.68$\pm$0.43\\
77.35$\pm$0.31\\
79.71$\pm$1.23\\
\bf{82.43$\pm$0.26}
} &\tabincell{c}{71.52$\pm$1.55\\
72.35$\pm$0.65\\
75.07$\pm$0.47\\
75.37$\pm$1.76\\
\bf{79.51$\pm$0.34}
} &\tabincell{c}{67.20$\pm$2.96\\
69.30$\pm$0.45\\
70.25$\pm$0.32\\
72.28$\pm$0.56\\
\bf{76.33$\pm$0.32}
} &\tabincell{c}{66.12$\pm$3.58\\
67.41$\pm$0.62\\
67.56$\pm$0.52\\
69.27$\pm$0.76\\
\bf{73.53$\pm$0.51}
} &\tabincell{c}{63.56$\pm$2.05\\
64.36$\pm$1.42\\
64.82$\pm$0.42\\
68.21$\pm$1.11\\
\bf{72.12$\pm$0.43}
}\\\hline
    5&\tabincell{c}{MeanMap \\ InvCal\\Alter-$\propto$SVM\\ Conv-$\propto$SVM\\ TL-LLP}  &\tabincell{c}{68.12$\pm$0.27\\
70.14$\pm$2.58\\
\bf{74.03$\pm$2.33}\\
70.44$\pm$0.72\\
73.98$\pm$0.63\\
} &\tabincell{c}{66.32$\pm$1.20\\
68.14$\pm$0.66\\
71.74$\pm$3.15\\
68.25$\pm$0.56\\
\bf{73.07$\pm$0.43}
} &\tabincell{c}{63.37$\pm$1.26\\
65.59$\pm$1.23\\
67.03$\pm$2.83\\
64.18$\pm$1.28\\
\bf{70.31$\pm$0.84}
} &\tabincell{c}{61.81$\pm$1.05\\
64.81$\pm$1.47\\
64.47$\pm$2.70\\
61.19$\pm$1.11\\
\bf{66.95$\pm$1.21}
} &\tabincell{c}{60.80$\pm$1.53\\
61.06$\pm$3.66\\
63.03$\pm$3.57\\
58.59$\pm$1.73\\
\bf{64.16$\pm$1.45}
}&\tabincell{c}{55.83$\pm$2.47\\
59.91$\pm$3.79\\
60.76$\pm$3.41\\
57.82$\pm$0.48\\
\bf{62.08$\pm$0.40}
}\\\hline
	6&\tabincell{c}{MeanMap \\ InvCal\\Alter-$\propto$SVM\\ Conv-$\propto$SVM\\ TL-LLP}
&\tabincell{c}{58.25$\pm$0.33\\
58.26$\pm$0.41\\
59.34$\pm$0.36\\
60.24$\pm$0.26\\
\bf{64.72$\pm$0.26}
} &\tabincell{c}{56.72$\pm$0.49\\
58.10$\pm$0.52\\
58.47$\pm$0.56\\
59.02$\pm$0.66\\
\bf{61.74$\pm$0.50}
} &\tabincell{c}{55.84$\pm$1.23\\
56.02$\pm$1.26\\
58.19$\pm$1.32\\
57.26$\pm$1.23\\
\bf{60.81$\pm$0.68}
} &\tabincell{c}{54.60$\pm$0.65\\
54.08$\pm$1.26\\
54.35$\pm$1.17\\
55.31$\pm$1.14\\
\bf{58.24$\pm$0.43}
} &\tabincell{c}{54.67$\pm$1.06\\
53.30$\pm$1.29\\
54.89$\pm$1.52\\
52.12$\pm$1.69\\
\bf{56.01$\pm$1.01}
} &\tabincell{c}{51.78$\pm$1.72\\
52.00$\pm$2.12\\
\bf{53.23$\pm$1.30}\\
51.07$\pm$1.05\\
53.21$\pm$0.89
}\\\hline
7&\tabincell{c}{MeanMap \\ InvCal\\Alter-$\propto$SVM\\ Conv-$\propto$SVM\\ TL-LLP}
&\tabincell{c}{75.46$\pm$1.33\\
76.25$\pm$1.45\\
78.24$\pm$1.06\\
76.12$\pm$0.66\\
\bf{81.87$\pm$0.64}
}&\tabincell{c}{71.71$\pm$1.25\\
73.87$\pm$0.72\\
78.02$\pm$0.83\\
74.08$\pm$1.16\\
\bf{80.02$\pm$0.64}
}&\tabincell{c}{68.89$\pm$1.54\\
69.27$\pm$2.16\\
76.87$\pm$0.72\\
71.38$\pm$1.94\\
\bf{79.03$\pm$0.52}
}&\tabincell{c}{67.64$\pm$1.63\\
65.58$\pm$2.03\\
74.72$\pm$1.34\\
70.91$\pm$1.35\\
\bf{77.63$\pm$1.12}
}&\tabincell{c}{68.20$\pm$1.46\\
65.50$\pm$3.28\\
70.30$\pm$1.48\\
68.77$\pm$2.94\\
\bf{76.84$\pm$1.33}
}&\tabincell{c}{67.83$\pm$1.73\\
63.98$\pm$2.22\\
68.43$\pm$0.96\\
72.66$\pm$1.23\\
\bf{76.21$\pm$0.73}
}\\\hline
8&\tabincell{c}{MeanMap \\ InvCal\\Alter-$\propto$SVM\\ Conv-$\propto$SVM\\ TL-LLP}
&\tabincell{c}{73.21$\pm$1.04\\
73.82$\pm$1.01\\
75.31$\pm$0.66\\
77.27$\pm$1.06\\
\bf{83.98$\pm$0.60}
}&\tabincell{c}{70.06$\pm$1.68\\
70.34$\pm$1.25\\
73.02$\pm$1.46\\
75.49$\pm$1.36\\
\bf{81.89$\pm$1.02}
}&\tabincell{c}{65.69$\pm$2.12\\
67.30$\pm$1.78\\
72.80$\pm$1.28\\
72.72$\pm$0.95\\
\bf{78.92$\pm$0.64}
}&\tabincell{c}{66.93$\pm$1.36\\
66.17$\pm$2.13\\
71.25$\pm$1.54\\
70.87$\pm$1.59\\
\bf{76.03$\pm$0.78}
}&\tabincell{c}{66.27$\pm$2.56\\
62.57$\pm$2.28\\
68.45$\pm$1.45\\
69.55$\pm$2.04\\
\bf{74.88$\pm$1.45}
}&\tabincell{c}{61.37$\pm$2.23\\
62.85$\pm$3.42\\
67.18$\pm$1.32\\
68.18$\pm$1.54\\
\bf{72.94$\pm$1.30}
}\\\hline
9&\tabincell{c}{MeanMap \\ InvCal\\Alter-$\propto$SVM\\ Conv-$\propto$SVM\\ TL-LLP}
&\tabincell{c}{70.47$\pm$0.38\\
75.74$\pm$3.10\\
77.51$\pm$0.61\\
75.39$\pm$0.62\\
\bf{80.25$\pm$0.41}
}&\tabincell{c}{70.02$\pm$0.77\\
74.69$\pm$0.56\\
72.17$\pm$0.48\\
74.63$\pm$0.55\\
\bf{78.48$\pm$0.32}
}&\tabincell{c}{68.76$\pm$1.21\\
72.80$\pm$0.70\\
74.15$\pm$0.84\\
71.71$\pm$0.45\\
\bf{76.36$\pm$0.30}
}&\tabincell{c}{66.37$\pm$0.52\\
70.58$\pm$0.48\\
\bf{72.78$\pm$0.45}\\
69.82$\pm$1.12\\
72.62$\pm0.42$
}&\tabincell{c}{65.25$\pm1.05$\\
67.60$\pm$1.56\\
66.53$\pm$1.05\\
65.46$\pm$1.43\\
\bf{70.43$\pm$0.88}
}&\tabincell{c}{63.10$\pm$2.28\\
65.35$\pm$1.62\\
66.81$\pm$0.62\\
65.98$\pm$1.32\\
\bf{68.42$\pm$0.57}
}\\\hline
\end{tabular}}
\end{table}
\subsection{Experimental Results}
In this section, we present the experimental results. Table \ref{tab:acc} shows the average accuracy and standard deviation of the methods under the bag size from 2 to 64. From the Table, we can know that the proposed TL-LLP method outperforms the MeanMap, InvCal, Alter-$\propto$SVM and Conv-$\propto$SVM methods on most data sets. This is because MeanMap, InvCal, Alter-$\propto$SVM and Conv-$\propto$SVM methods are proposed for single task, and they only build classifier using the target task, however, the proposed method can transfer knowledge from the source task to the target task to help the target task build a more predictive classifier, thus, the proposed TL-LLP can deliver higher performance for most of the data sets. In addition, we observe even for the dataset 5 with bag number 2, dataset 6 with bag number 64, and dataset 9 with bag number 16, the TL-LLP does not obtain the highest accuracy, its performance is close to the highest accuracy. In addition, Table \ref{tab:acc} also demonstrates the standard derivation of each method on the datasets. We can observe that the standard derivation of the proposed method is less than other methods on most of the datasets, which implies that the proposed TL-LLP method can deliver a stabler performance than other LLP methods.

Furthermore, Figure \ref{fig:meac} shows the mean accuracy on data sets under bag size from 2 to 64. From Figure \ref{fig:meac}, we can see that the average performance of TL-LLP is higher than MeanMap, InvCal, Alter-$\propto$SVM and Conv-$\propto$SVM methods. We can conclude that the proposed TL-LLP method can well solve LLP problems.
\begin{figure}[ht]
\begin{minipage}[t]{0.32\textwidth}
\centering
\includegraphics[width=5.7cm,height=3.5cm]{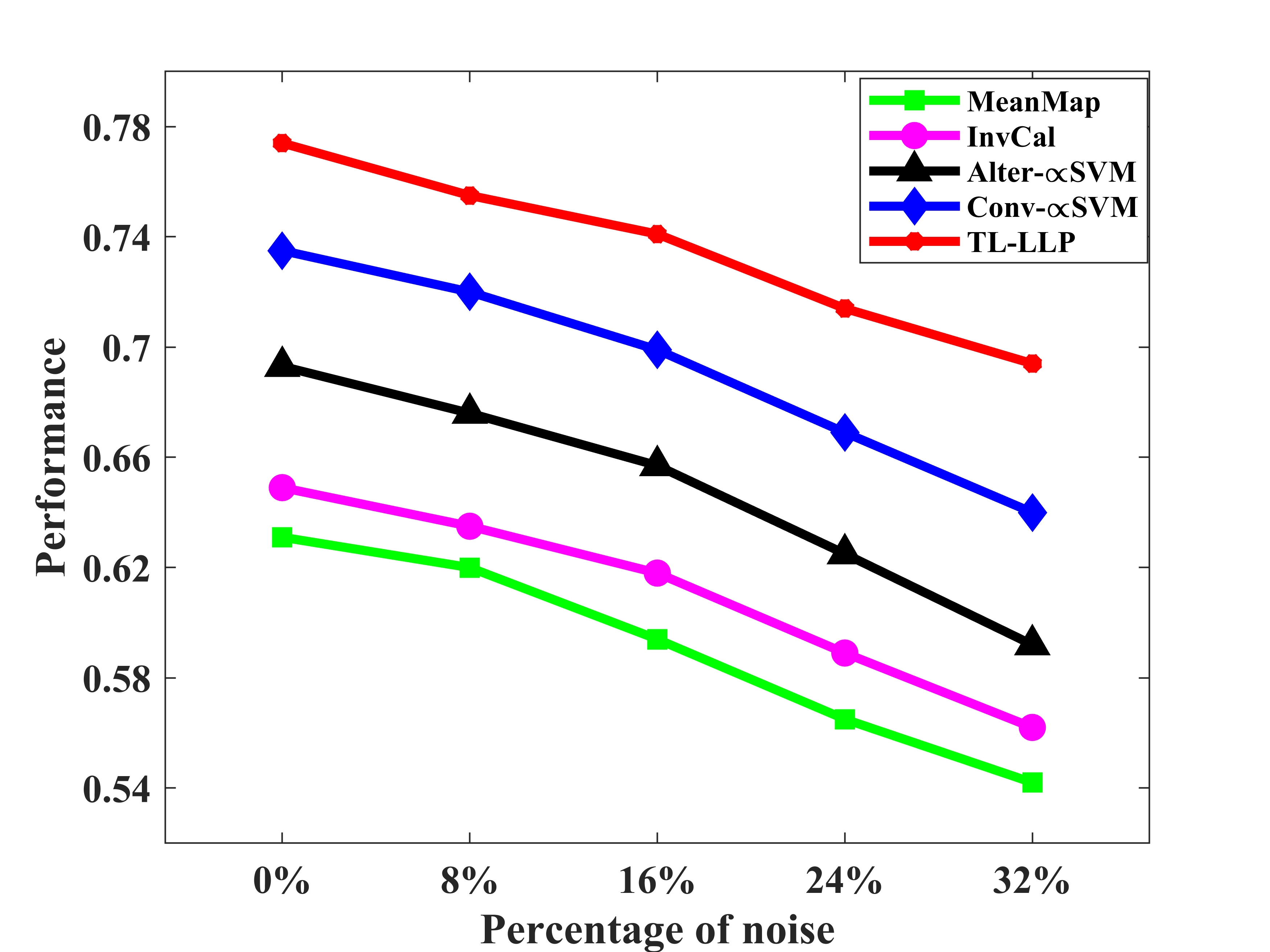}
\subcaption*{Dataset 1}
\end{minipage}
\begin{minipage}[t]{0.32\textwidth}
\centering
\includegraphics[width=5.7cm,height=3.5cm]{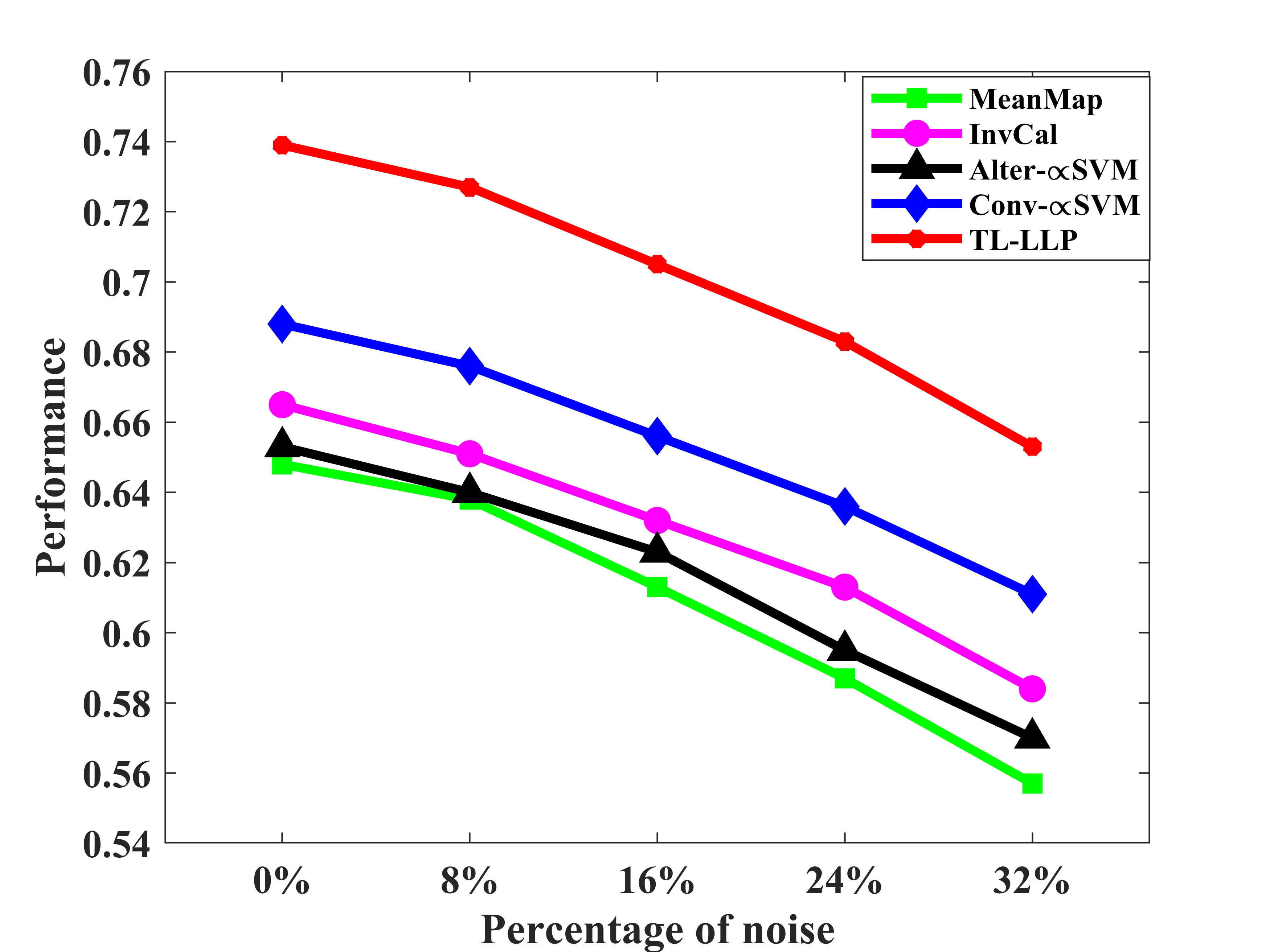}
\subcaption*{Dataset 2}
\end{minipage}
\begin{minipage}[t]{0.32\textwidth}
\centering
\includegraphics[width=5.7cm,height=3.5cm]{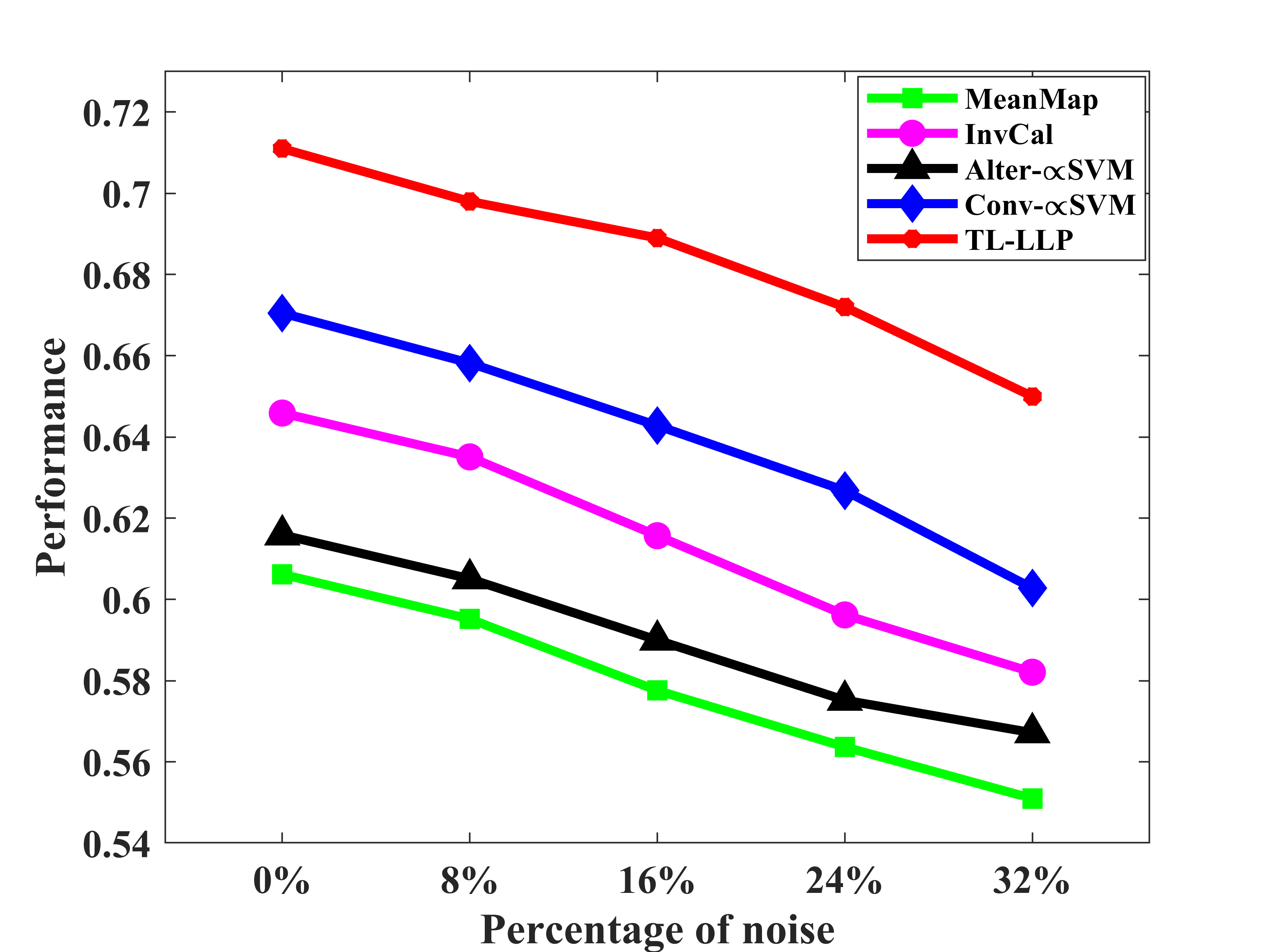}
\subcaption*{Dataset 3}
\end{minipage}
\begin{minipage}[t]{0.32\textwidth}
\centering
\includegraphics[width=5.7cm,height=3.5cm]{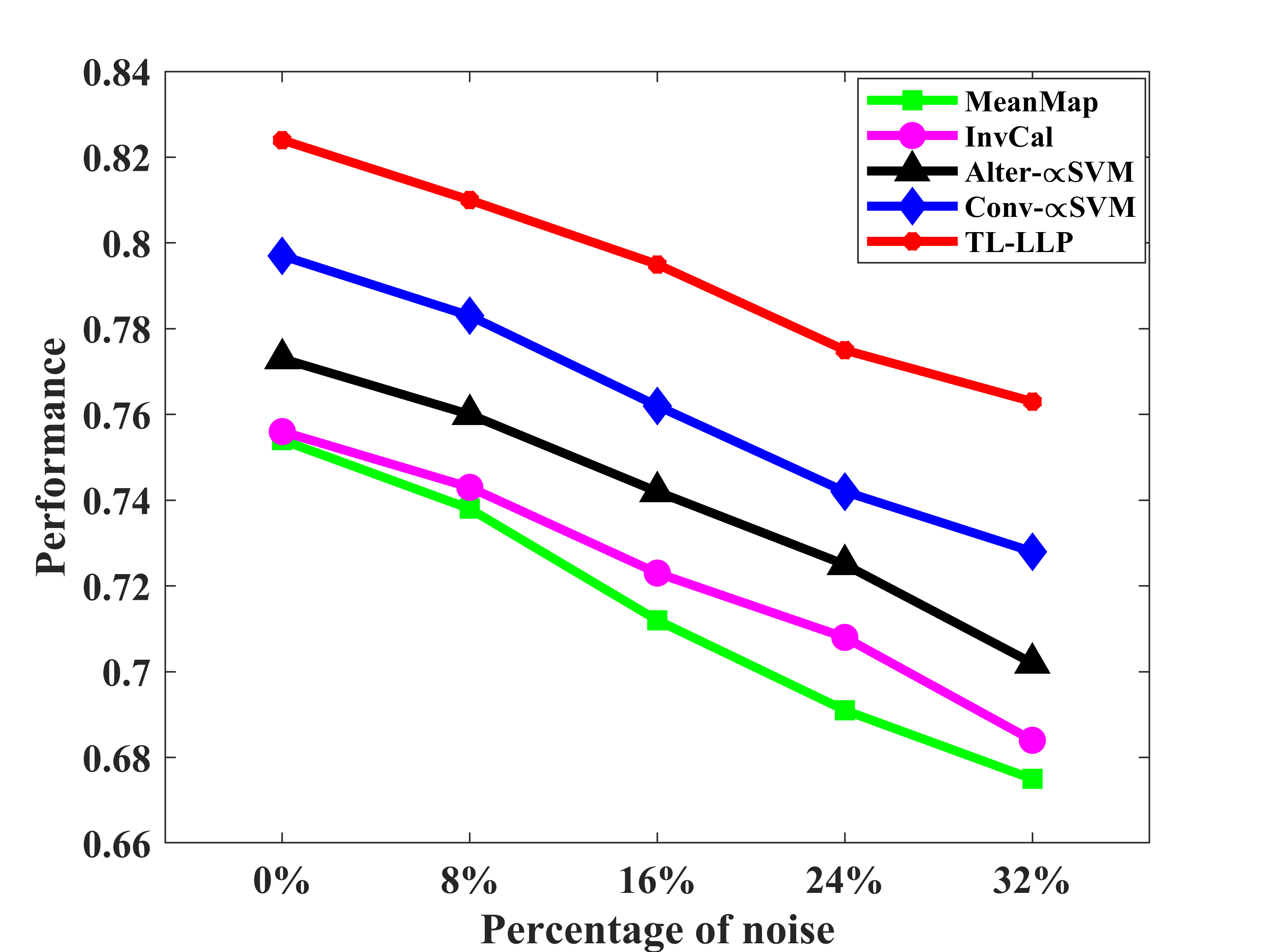}
\subcaption*{Dataset 4}
\end{minipage}
\begin{minipage}[t]{0.32\textwidth}
\centering
\includegraphics[width=5.7cm,height=3.5cm]{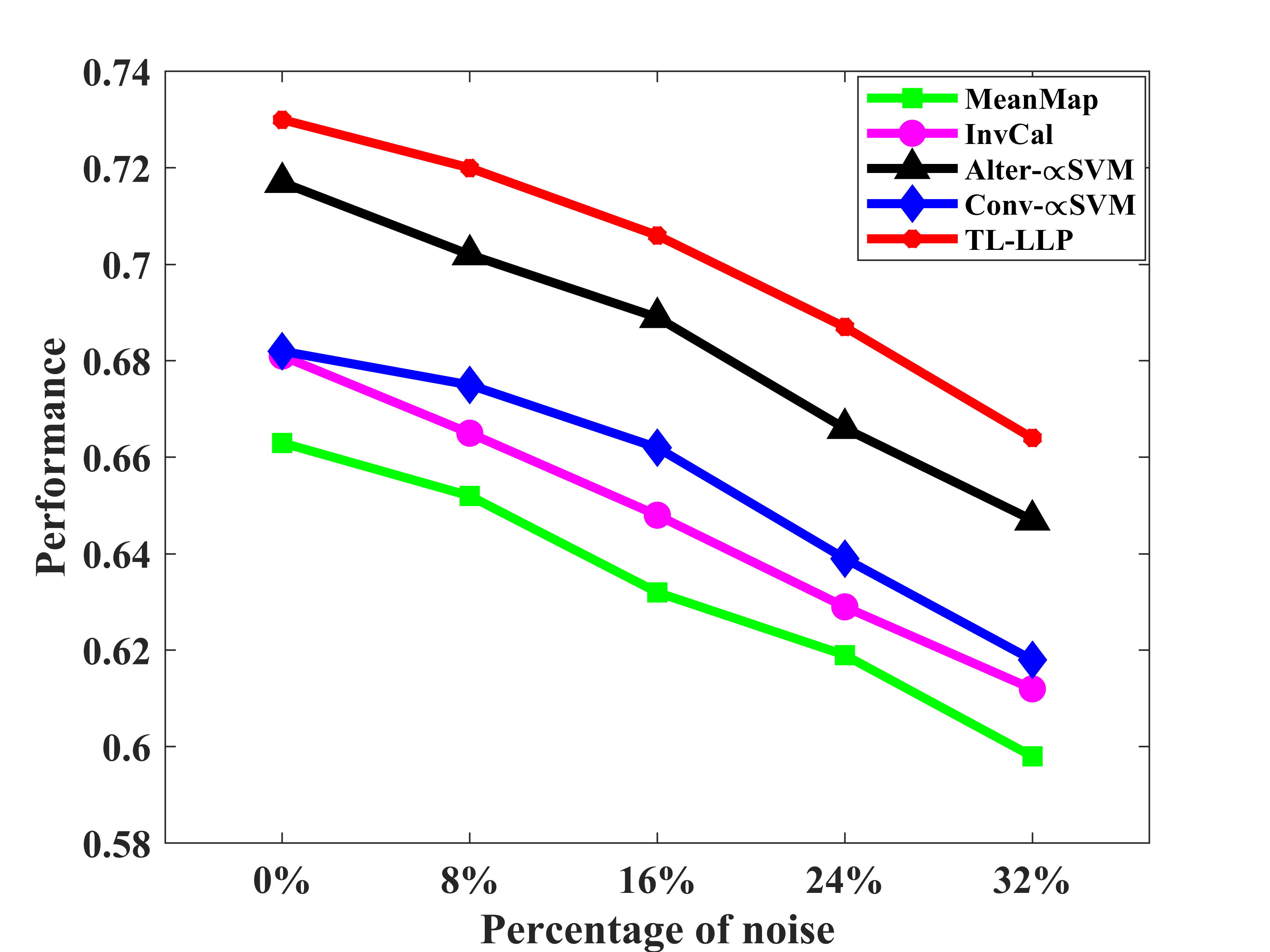}
\subcaption*{Dataset 5}
\end{minipage}
\begin{minipage}[t]{0.32\textwidth}
\centering
\includegraphics[width=5.7cm,height=3.5cm]{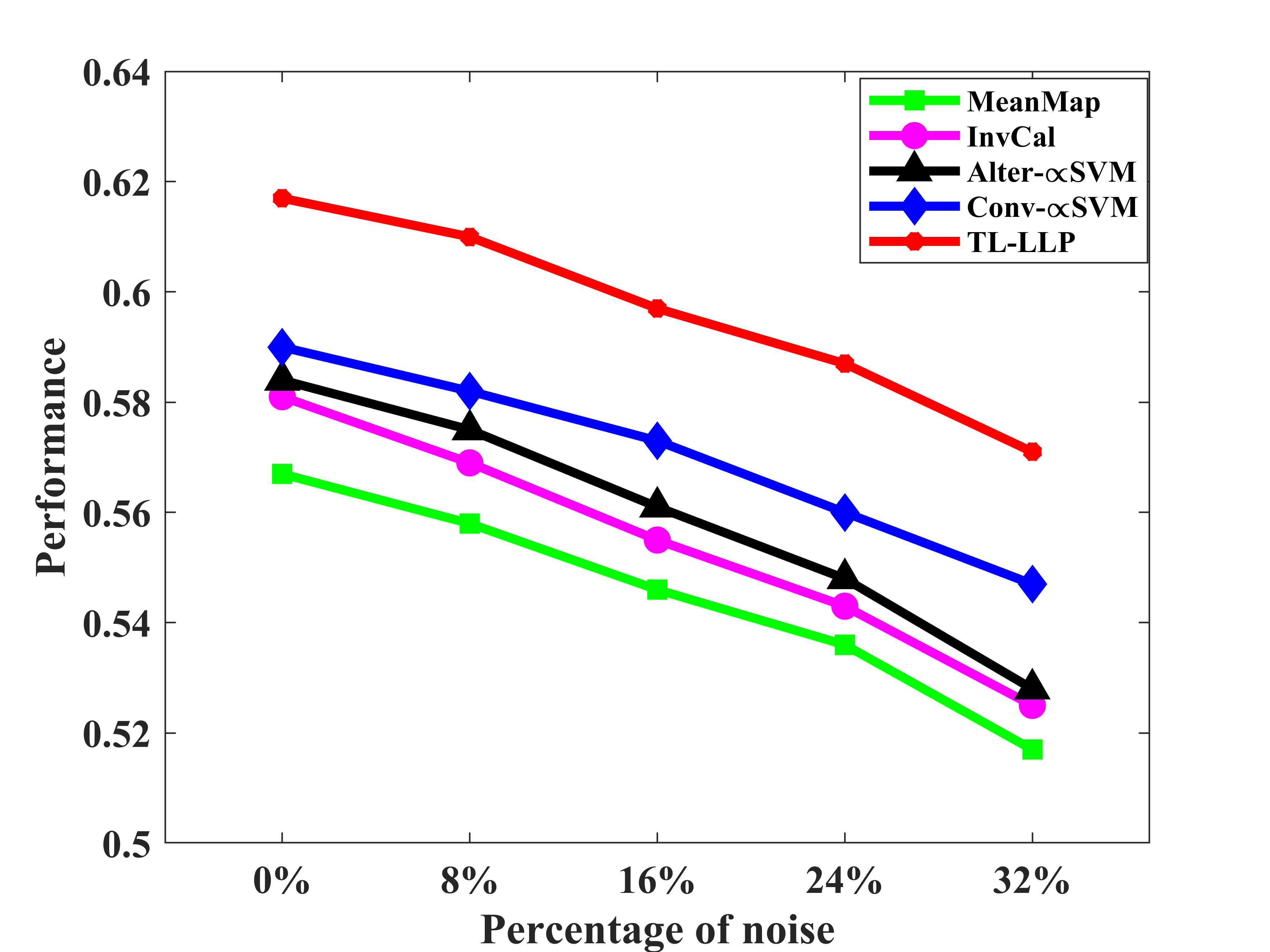}
\subcaption*{Dataset 6}
\end{minipage}
\begin{minipage}[t]{0.32\textwidth}
\centering
\includegraphics[width=5.7cm,height=3.5cm]{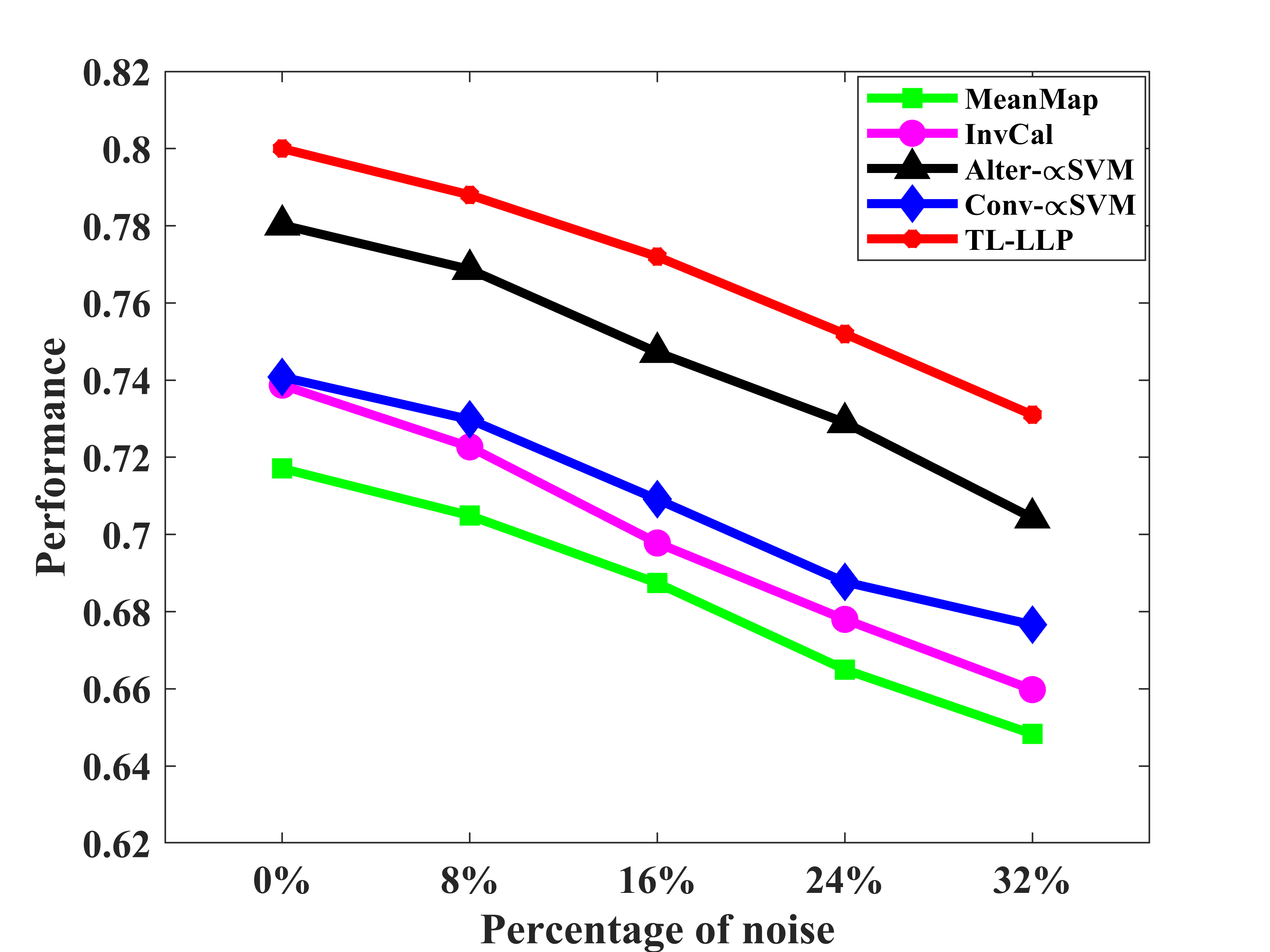}
\subcaption*{Dataset 7}
\end{minipage}
\begin{minipage}[t]{0.32\textwidth}
\centering
\includegraphics[width=5.7cm,height=3.5cm]{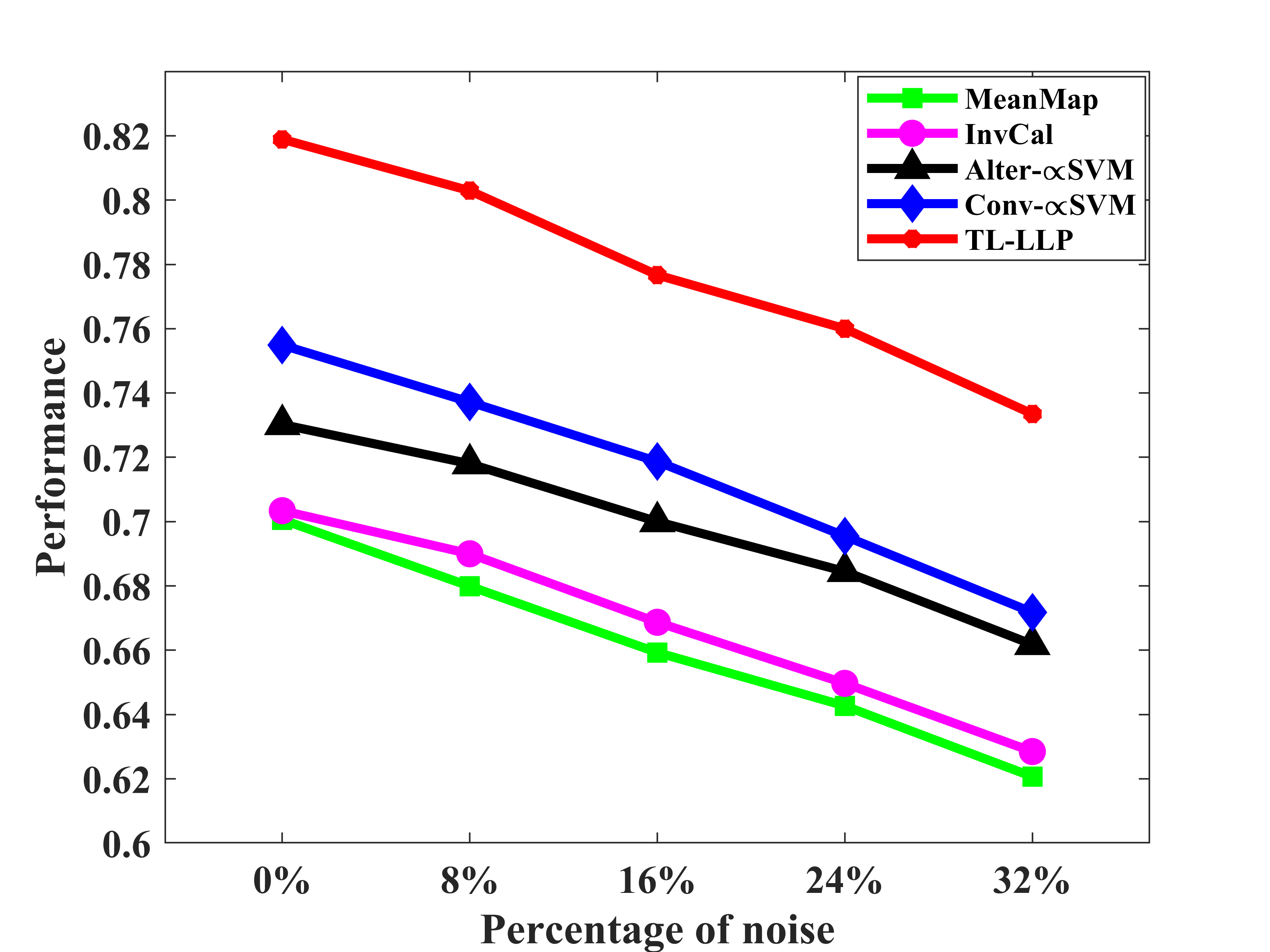}
\subcaption*{Dataset 8}
\end{minipage}
\begin{minipage}[t]{0.32\textwidth}
\centering
\includegraphics[width=5.7cm,height=3.5cm]{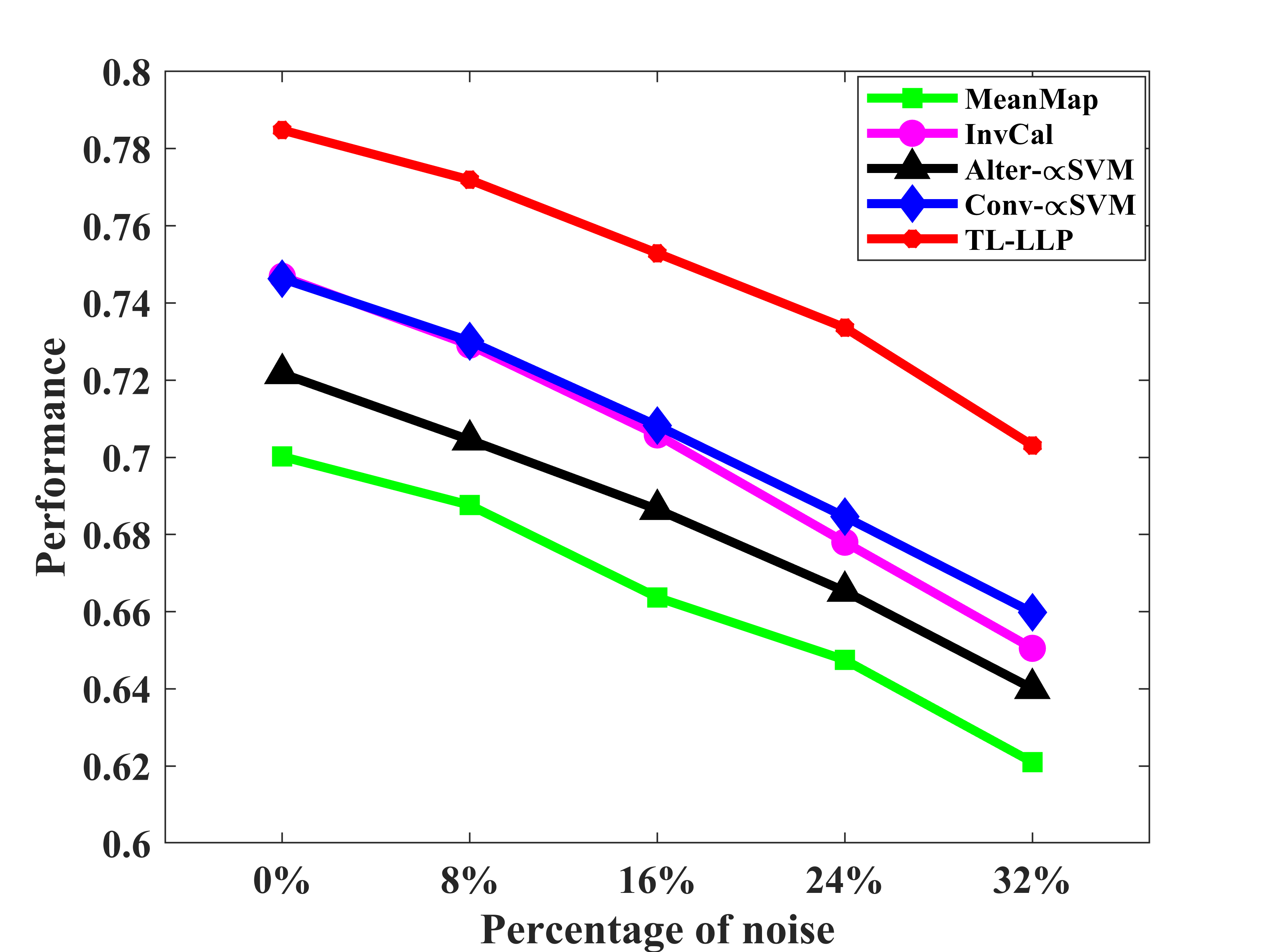}
\subcaption*{Dataset 9}
\end{minipage}
\caption{Accuracy comparison among different percents of data corrupted by noise for different data sets}
\label{fig:sensitivity}
\end{figure}
\subsection{Sensitivity to Input Data Noise}
To compare the performance of five algorithms with regards to dealing with uncertain data, we add the noise into the source and target tasks data and conduct experiments to investigate the sensitivity of the five algorithms to the noise. Following the method used in \cite{aggarwal2008outlier,liu2014efficient}, we generate the noise using a Gaussian distribution with zero mean and standard deviation determined as follows.

For each data set, we first calculate the standard deviation $\sigma_i^0$ of the entire data along the $i$th dimension, which is to model the difference in noises on different dimensions. Then we obtain the standard deviation of the Gaussian noise $\sigma_i$ randomly from the range [0,2$\cdot\sigma_i^0$] to add noises for $i$th dimension.  By doing this, a data example $x_j$ is added with noises, which can be presented as a vector
\begin{align}
\sigma^{x_j}=[\sigma_1^{x_j},\sigma_2^{x_j},...,\sigma_{d-1}^{x_j},\sigma_d^{x_j}]
\end{align}

Here, $d$ denotes the number of dimensions for a data example $x_j$ , and $\sigma_i^{x_j},i = 1,\cdots,d$ represents the noise added into the $i$th dimension of the data example. Figure \ref{fig:noise} declares the basic idea of the method of adding noise to data examples. In this figure, {\bfseries x} is the original data example, {\bfseries v} is a noise vector, the new data example with added noises is represented by $\bf{x^v}$, which has some deviations from the original example {\bfseries x}.

In our experiment, we make the percentage of data added noises increases from 0 to 32\%, and apply these noisy datasets to MeanMap, InvCal, Alter-$\propto$SVM, Conv-$\propto$SVM and TL-LLP methods. Figure \ref{fig:sensitivity} illustrates the accuracy values obtained for the five methods related to different percentages of data corrupted by noise.  We can discover that, as more noise is added to the data, the accuracy values of the five methods decrease. This is because as the noise percentage increases, the positive class potentially becomes less distinguishable from the negative class. However, it is easily to see that our method TL-LLP can still consistently yield higher accuracy than MeanMap, InvCal, Alter-$\propto$SVM and Conv-$\propto$SVM, which implies that the proposed TL-LLP method can reduces the effect of noises in the data.
\subsection{Running time analysis}
\begin{figure}[tbp]
\centering
\includegraphics[width=10cm]{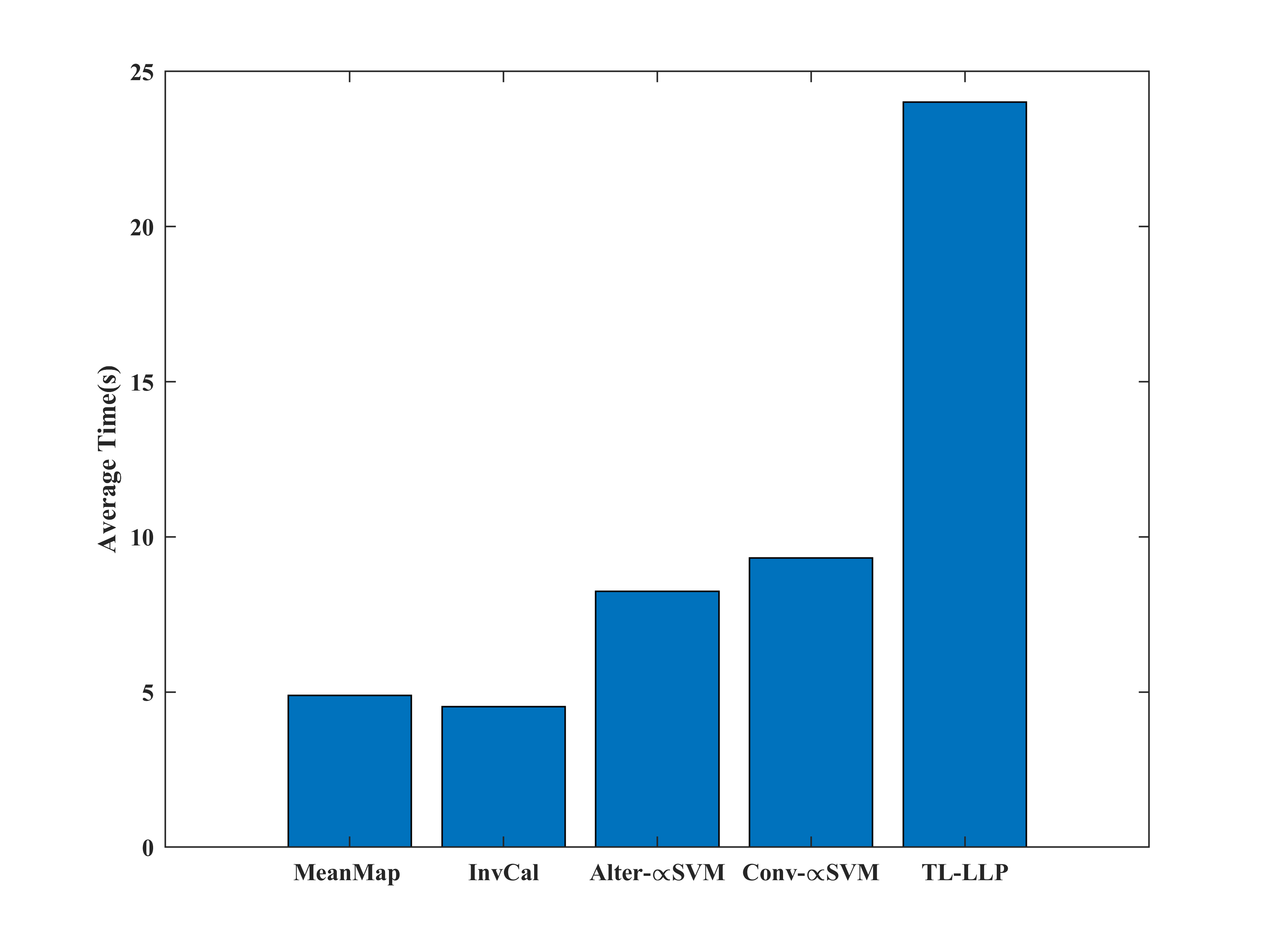}
\caption{Average computational times on different methods}
\label{fig:times}
\end{figure}
Figure \ref{fig:times} presents the average computational time of five methods on different data sets. From this figure, we can see that the computational time required by TL-LLP is more than other methods. This is because that the MeanMap, InvCal, Alter-$\propto$SVM and Conv-$\propto$SVM methods are single task model,  which is only trained on target task, and the TL-LLP method is two tasks model, which involves both source task dataset and target task dataset into learning. Thus, the proposed TL-LLP method requires more running time compared with other methods. In addition, we observe that Alter-$\propto$SVM and Conv-$\propto$SVM take more time than MeanMap and InvCal methods, since $\propto$SVM model needs to solve a quadratic optimization in the learning.
\section{Conclusions and Future Work}
This paper addresses the problem of transfer learning-based learning with label proportion. To assist the target task to learn a model for prediction, this paper has proposed a transfer learning model for learning a classifier from label proportions. Our proposed approach transfers knowledge from the source task to the target task. We then convert the objective model into the Dual problem using Lagrange method. Extensive experiments have been conducted to investigate the performance of our proposed approach, and the experiments show that the proposed method outperforms the existing LLP methods.

In the future, we would like to apply the proposed method on the data stream environments.
\section{APPENDIX}
In this section, we will present the detailed derivation of Lemmas.
\subsection{Proof of Lemma 1}
 In the optimization problem (\ref{equ:problem}), if fix each $\triangle\bm{x_i}$ as a small value and $||\triangle\bm{x_i}|| \leq \delta_i$, the constraint $\triangle\bm{x_i} \leq \delta_i$ is useless and will not have any impact on this optimization problem, since $||\triangle\bm{x_i}||$ is already less than or equal $\delta_i$. Thus, we can delete the constraint from problem (\ref{equ:problem}), and then use the optimization problem (\ref{equ:prime}) to replace optimization (\ref{equ:problem}).
\subsection{Proof of Lemma 2}
To solve the primal problem (\ref{equ:prime}), let $\overline{\bm{x}}_i=\bm{x_i}+\triangle\overline{\bm{x}}_i$ and  introduce multipliers $\alpha$ and $\gamma$ build the Lagrange function as
    \begin{align}\label{equ:L}
    \allowdisplaybreaks[4]
\setlength{\abovedisplayskip}{3pt}
\setlength{\belowdisplayskip}{3pt}
    L&=\frac{1}{2}||\bm{w}_0||{^2}+\frac{\lambda_1}{2}||\bm{v}_1||{^2}+\frac{\lambda_2}{2}||\bm{v}_2||{^2}\notag\\
    &+C_1\sum\limits_{i=1}^{t_1}(\xi_{1i}+\xi_{1i}^*)+C_2\sum\limits_{m=1}^{t_2}(\xi_{2m}+\xi_{2m}^*)\notag\\
    &-\sum\limits_{i=1}^{t_1}\alpha_{1i}(\xi_{1i}+\varepsilon_{1i}+y_{1i}-\frac{1}{|B_i^{T_1}|}\sum\limits_{j\in B_i^{T_1}}(\bm{w}_1^{T}\overline{\bm{x}}_{1j}+b_1))\notag\\
    &-\sum\limits_{i=1}^{t_1}\alpha_{1i}^*(\xi_{1i}^*+\varepsilon_{1i}+\frac{1}{|B_i^{T_1}|}\sum\limits_{j\in B_i^{T_1}}(\bm{w}_1^{T}\overline{\bm{x}}_{1j}+b_1)-y_{1i})\notag\\
    &-\sum\limits_{m=1}^{t_2}\alpha_{2m}(\xi_{2m}+\varepsilon_{2m}+y_{2m}-\frac{1}{|B_m^{T_2}|}\sum\limits_{n\in B_m^{T_2}}(\bm{w}_2^{T}\overline{\bm{x}}_{2n}+b_2))\notag\\
    &-\sum\limits_{m=1}^{t_2}\alpha_{2m}^*(\xi_{2j}^*+\varepsilon_{2m}+\frac{1}{|B_m^{T_2}|}\sum\limits_{n\in B_m^{T_2}}(\bm{w}_2^{T}\overline{\bm{x}}_{2n}+b_2)-y_{2m})\notag\\
    &-\sum\limits_{i=1}^{t_1}(\gamma_{1i}\xi_{1i}+\gamma_{1i}^*\xi_{1i}^*)-\sum\limits_{m=1}^{t_2}(\gamma_{2m}\xi_{2m}+\gamma_{2m}^*\xi_{2m}^*).
    \end{align}

To minimize L, we derive partial derivatives $\bm{w}_0,\bm{v}_1,\bm{v}_2,b_1,b_2$ and $\xi(*)$ equal to zero, we have
    $\nabla_{\bm{w}}(L)=0,\quad\nabla_{\bm{v}_1}(L)=0,\quad\nabla_{\bm{v}_2}(L)=0,\quad\nabla_b(L)=0,\quad\nabla_{\xi(*)}(L)=0$, and we get
\begin{align}
\label{equ:w}&\overline{\bm{w}}_0=\sum\limits_{i=1}^{t_1}\frac{(\alpha_{1i}^*-\alpha_{1i})}{|B_i^{T_1}|}\sum\limits_{j\in B_i^{T_1}}\overline{\bm{x}}_{1j}+\sum\limits_{m=1}^{t_2}\frac{(\alpha_{2m}^*-\alpha_{2m})}{|B_m^{T_2}|}\sum\limits_{n\in B_m^t}\overline{\bm{x}}_{2n}\\
\label{equ:v1}&\overline{\bm{v}}_1=\frac{1}{\lambda_1}\sum\limits_{i=1}^{t_1}(\alpha_{1i}^*-\alpha_{1i})\frac{1}{|B_i^{T_1}|}\sum\limits_{j\in B_i^{T_1} }\overline{\bm{x}}_{1j}\\
\label{equ:v2}&\overline{\bm{v}}_2=\frac{1}{\lambda_2}\sum\limits_{m=1}^{t_2}(\alpha_{2m}^*-\alpha_{2m})\frac{1}{|B_m^{T_2}|}\sum\limits_{n\in B_m^{T_2}}\overline{\bm{x}}_{2n}\\
    \label{equ:a}&\sum\limits_{i=1}^{t_1}(\alpha_{1i}-\alpha_{1i}^*)=0\\
    &\sum\limits_{m=1}^{t_2}(\alpha_{2m}-\alpha_{2m}^*)=0\\
    &C_1-\alpha_{1i}^{(*)}-\gamma_{1i}^{(*)}=0(i=1,2,\cdots,t_1)\\
    \label{equ:c}&C_2-\alpha_{2m}^{(*)}-\gamma_{2m}^{(*)}=0(m=1,2,\cdots,t_2)
\end{align}
    Substituting the equation (\ref{equ:w}-\ref{equ:c}) into equation (\ref{equ:L}) to obtain the dual form (\ref{equ:dual}) of the problem (\ref{equ:prime}).
\subsection{Proof of Lemma 3}
If $\bm{w}_0, \bm{v}_1, \bm{v}_2, b_1$ and $b_2$ is fixed to be $\overline{\bm{w}}_0, \overline{\bm{v}}_1, \overline{\bm{v}}_2, \overline{b}_1$ and $\overline{b}_2$  in problem (\ref{equ:problem}), the optimization of problem (\ref{equ:problem}) over $\triangle\overline{\bm{x}}$ equals to minimization of $\sum\limits_{i=1}^{t_1}(\xi_{1i}+\xi_{1i}^*)+\sum\limits_{m=1}^{t_2}(\xi_{2m}+\xi_{2m}^*)$ over each $\triangle\overline{\bm{x}}$.

We assume each noise vector $\triangle\bm{x}$ just corrupts sample $\bm{x}_i$ and will not affects other instances. Consequently, for the source task, $\triangle\bm{x}_{1i}$ have impact on $\xi_{1i}$ or $\xi_{1i}^*$. The optimization of $\sum\limits_{i=1}^{t_1}(\xi_{1i}+\xi_{1i}^*)$ can be divided to minimize each $\xi_{1i}$ and $\xi_{1i}^*, i=1,2\ldots,t_1$. From the first and second constraint of problem (\ref{equ:problem}), if $|f_1(x_i)-y_{1i}|\leq\varepsilon_{1i}$, $\xi_{1i}=\xi_{1i}^*=0$; if $f_1(x_i)-y_{1i}>\varepsilon_{1i}$, $\xi_{1i}^*=0$ and $\triangle\bm{x}_{1i}$ only have impact on $\xi_{1i}$; if $y_{1i}-f_1(x_i)>\varepsilon_{1i}$, $\xi_{1i}=0$ and $\triangle\bm{x}_{1i}$ only have impact on $\xi_{1i}^*$.

For minimization of each $\xi_{1i}$, from the first constraint of problem (\ref{equ:problem}), we have
\begin{equation}\label{equ:xi1}
\setlength{\abovedisplayskip}{3pt}
\setlength{\belowdisplayskip}{3pt}
\begin{split}
\xi_{1i}=max(0,&\frac{1+\lambda_1}{\lambda_1}\sum\limits_{j=1}^{t_1}\frac{(\alpha_{1j}^*-\alpha_{1j})}{|B_j^{T_1}|}\sum\limits_{k\in B_j^{T_1}}K(\bm{x}_{1k}+\triangle\bm{x}_{1k},\bm{x}_i+\triangle\bm{x}_i)\\
&+\sum\limits_{m=1}^{t_2}\frac{(\alpha_{2m}^*-\alpha_{2m})}{|B_m^{T_2}|}\sum\limits_{n\in B_m^{T_2}}K(\bm{x}_{2n}+\triangle\bm{x}_{2n},\bm{x}_i+\triangle\bm{x}_i)+b_1-y_{1i}-\varepsilon_{1i})
\end{split}
\end{equation}
Using Taylor expansjion \cite{kline1998calculus}, we have
\begin{align}
\allowdisplaybreaks[4]
\setlength{\abovedisplayskip}{3pt}
\setlength{\belowdisplayskip}{3pt}
\sum\limits_{j=1}^{t_1}&\frac{(\alpha_{1j}^*-\alpha_{1j})}{|B_j^{T_1}|}\sum\limits_{k\in B_j^{T_1}}K(\bm{x}_{1k}+\triangle\bm{x}_{1k},\bm{x}_i+\triangle\bm{x}_i)\notag\\
&=\sum\limits_{j=1}^{t_1}\frac{(\alpha_{1j}^*-\alpha_{1j})}{|B_j^{T_1}|}\sum\limits_{k\in B_j^{T_1}}K(\bm{x}_{1k}+\triangle\bm{x}_{1k},\bm{x}_i)\notag\\
&+\triangle\bm{x}_i^T\sum\limits_{j=1}^{t_1}\frac{(\alpha_{1j}^*-\alpha_{1j})}{|B_j^{T_1}|}\sum\limits_{k\in B_j^{T_1}}K^{\prime}(\bm{x}_{1k}+\triangle\bm{x}_{1k},\bm{x}_i)
\end{align}
\begin{equation}
\setlength{\abovedisplayskip}{3pt}
\setlength{\belowdisplayskip}{3pt}
\begin{split}
\sum\limits_{m=1}^{t_2}&\frac{(\alpha_{2m}^*-\alpha_{2m})}{|B_m^{T_2}|}\sum\limits_{n\in B_m^{T_2}}K(\bm{x}_{2n}+\triangle\bm{x}_{2n},\bm{x}_i+\triangle\bm{x}_i)\\
&=\sum\limits_{m=1}^{t_2}\frac{(\alpha_{2m}^*-\alpha_{2m})}{|B_m^{T_2}|}\sum\limits_{n\in B_m^{T_2}}K(\bm{x}_{2n}+\triangle\bm{x}_{2n},\bm{x}_i)\\
&+\triangle\bm{x}_i^T\sum\limits_{m=1}^{t_2}\frac{(\alpha_{2m}^*-\alpha_{2m})}{|B_m^{T_2}|}\sum\limits_{n\in B_m^{T_2}}K^{\prime}(\bm{x}_{2n}+\triangle\bm{x}_{2n},\bm{x}_i)
\end{split}
\end{equation}
Let $\bm{u}_1=\frac{1+\lambda_1}{\lambda_1}\sum\limits_{j=1}^{t_1}\frac{(\alpha_{1j}^*-\alpha_{1j})}{|B_j^{T_1}|}\sum\limits_{k\in B_j^{T_1}}K^{\prime}(\bm{x}_{1k}+\triangle\bm{x}_{1k},\bm{x}_i)+\sum\limits_{m=1}^{t_2}\frac{(\alpha_{2m}^*-\alpha_{2m})}{|B_m^{T_2}|}\sum\limits_{n\in B_m^{T_2}}K^{\prime}(\bm{x}_{2n}+\triangle\bm{x}_{1n},\bm{x}_i)$ thus (\ref{equ:xi1}) equals to
\begin{equation}\label{equ:u}
\setlength{\abovedisplayskip}{3pt}
\setlength{\belowdisplayskip}{3pt}
\xi_{1i}=max(0,\frac{1+\lambda_1}{\lambda_1}\bm{w}_1^T\bm{x}_i+\bm{w}_2^T\bm{x}_i
+{\bm{u}_1}^T\triangle\bm{x}_i+b_1-y_{1i}-\varepsilon_{1i})
\end{equation}
From (\ref{equ:u}), it is seen that we can minimize $\xi_{1i}$ by maximizing   $-{\bm{u}_1}^T\triangle\bm{x}_i$. By using the Cauchy-Schwarz inequality \cite{dragomir2003survey}, it has
\begin{equation}
-||\bm{u}_1||\cdot||\triangle\bm{x}_i||\leq -\bm{u}_1\cdot\triangle\bm{x}_i \leq ||\bm{u}_1||\cdot||\triangle\bm{x}_i||
\end{equation}
The equality holds if and only if $\triangle\bm{x}_i=-c\bm{u}_1$, where $c$ is a  constant
number. Since $||\triangle\bm{x}_i||\leq \delta_{1i}$,  the optimal value of $\triangle\overline{\bm{x}}_i$ is
\begin{equation}
\triangle\overline{\bm{x}}_i= \delta_{1i}\frac{-\bm{u}_1}{||\bm{u}_1||}.
\end{equation}

Similar to the above operation, we can minimize $\xi_{1i}^*$ by maximizing   ${\bm{u}_1}^T\triangle\bm{x}_i$ when $y_{1i}-f_1(x_i)>\varepsilon_{1i}$, we then have optimal
\begin{equation}
\triangle\overline{\bm{x}}_i= \delta_{1i}\frac{\bm{u}_1}{||\bm{u}_1||}.
\end{equation}
For the examples $\bm{x}_{1i}$ in source task, the $\triangle\overline{\bm{x}}_{1i}$ is
\[ \triangle\overline{\bm{x}}_{1i} =
\begin{cases}
\delta_{1i}\frac{-\bm{u}_1}{||\bm{u}_1||} & \text{if } f_1(\bm{x}_{1i})-y_{1i}>\varepsilon_{1i},\\
0 & \text{if } |f_1(\bm{x}_{1i})-y_{1i}|<\varepsilon_{1i},\\
\delta_{1i}\frac{\bm{u}_1}{||\bm{u}_1||} & \text{if } y_{1i}-f_1(\bm{x}_{1i})>\varepsilon_{1i}.
\end{cases} \]
For the $\bm{x}_{2m}$ in target task, similar to the operations as $\bm{x}_{1i}$, the $\triangle\overline{\bm{x}}_{2m}$ is
\[ \triangle\overline{\bm{x}}_{2m} =
\begin{cases}
\delta_{1i}\frac{-\bm{u}_2}{||\bm{u}_2||} & \text{if } f_1(\bm{x}_{2m})-y_{2m}>\varepsilon_{2m},\\
0 & \text{if } |f_1(\bm{x}_{2m})-y_{2m}|<\varepsilon_{2m},\\
\delta_{1i}\frac{\bm{u}_2}{||\bm{u}_2||} & \text{if } y_{1i}-f_1(\bm{x}_{2m})>\varepsilon_{2m}.
\end{cases} \]
where it has
\begin{equation*}
\bm{u}_2=\sum\limits_{j=1}^{t_1}\frac{(\alpha_{1j}^*-\alpha_{1j})}{|B_j^{T_1}|}\sum\limits_{k\in B_j^{T_1}}K^{\prime}(\bm{x}_{1k}+\triangle\bm{x}_{1k},\bm{x}_i)+\frac{1+\lambda_2}{\lambda_2}\sum\limits_{m=1}^{t_2}\frac{(\alpha_{2m}^*-\alpha_{2m})}{|B_m^{T_2}|}\sum\limits_{n\in B_m^{T_2}}K^{\prime}(\bm{x}_{2n}+\triangle\bm{x}_{1n},\bm{x}_i)
\end{equation*}
\bibliographystyle{unsrt}
\bibliography{references}
\end{document}